\documentclass{ws-ijns}

\usepackage{amsmath}
\usepackage{float}
\usepackage{multirow}
\usepackage{soul}
\usepackage[table,xcdraw]{xcolor}
\usepackage{diagbox}
\usepackage{hyperref}
\usepackage{xurl}
\usepackage{breakurl}
\usepackage{array}

\usepackage{graphicx}
\usepackage[english]{babel}
\usepackage[super,sort,compress]{cite}

\begin{document}

\catchline{31}{3}{2021}{2130001}{}

\markboth{Pedro Lara-Benítez, Manuel Carranza-García and José C. Riquelme}{Experimental Review on Deep Learning for Time Series Forecasting}

\title{An Experimental Review on Deep Learning Architectures\\for Time Series Forecasting\footnote{When referring to this paper, please cite the published version: P. Lara-Benítez, M. Carranza-García and J. C. Riquelme, An experimental review on deep learning architectures for time series forecasting. \textit{International Journal of Neural Systems} 31(03),  2130001  (2021).  \url{https://doi.org/10.1142/S0129065721300011}}}

\author{Pedro Lara-Benítez\footnote{Corresponding author}}
\author{Manuel Carranza-García}
\author{José C. Riquelme}

\address{Division of Computer Science, University of Sevilla,\\ES-41012 Seville, Spain\\
E-mail: plbenitez@us.es\\
www.us.es}

\maketitle

\begin{abstract}
In recent years, deep learning techniques have outperformed traditional models in many machine learning tasks. Deep neural networks have successfully been applied to address time series forecasting problems, which is a very important topic in data mining. They have proved to be an effective solution given their capacity to automatically learn the temporal dependencies present in time series. However, selecting the most convenient type of deep neural network and its parametrization is a complex task that requires considerable expertise. Therefore, there is a need for deeper studies on the suitability of all existing architectures for different forecasting tasks. In this work, we face two main challenges: a comprehensive review of the latest works using deep learning for time series forecasting; and an experimental study comparing the performance of the most popular architectures. The comparison involves a thorough analysis of seven types of deep learning models in terms of accuracy and efficiency. We evaluate the rankings and distribution of results obtained with the proposed models under many different architecture configurations and training hyperparameters. The datasets used comprise more than 50000 time series divided into 12 different forecasting problems. By training more than 38000 models on these data, we provide the most extensive deep learning study for time series forecasting. Among all studied models, the results show that long short-term memory (LSTM) and convolutional networks (CNN) are the best alternatives, with LSTMs obtaining the most accurate forecasts. CNNs  achieve comparable performance with less variability of results under different parameter configurations, while also being more efficient.
\end{abstract}


\begin{multicols}{2}

\section{Introduction}
Time series forecasting (TSF) plays a key role in a wide range of real-life problems that have a temporal component. Predicting the future through TSF is an essential research topic in many fields such as the weather \cite{Zhao:2016}, energy consumption \cite{Deb:2017}, financial indices \cite{Chen:2015}, retail sales \cite{rafiei:2015}, medical monitoring \cite{Tuarob:2017}, anomaly detection \cite{Ashok:2018}, traffic prediction \cite{Zhang:2019}
, etc. The unique characteristics of time series data, in which observations have a chronological order, often make their analysis a challenging task. Given its complexity, TSF is an area of paramount importance in data mining. TSF models need to take into account several issues such as trend and seasonal variations of the series and the correlation between observed values that are close in time. Therefore, over the last decades, researchers have placed their efforts on developing specialized models that can capture the underlying patterns of time series, so that they can be extrapolated to the future effectively.

In recent times, the use of deep learning (DL) techniques has become the most popular approach for many machine learning problems, including TSF \cite{Schmidhuber:2015}. Unlike classical statistical-based models that can only model linear relationships in data, deep neural networks have shown a great potential to map complex non-linear feature interactions \cite{Sebastian:2019}. Modern neural systems base their success in their deep structure, stacking several layers and densely connecting a large number of neurons. The increase in computing capacity during the last years has allowed creating deeper models, which has significantly improved their learning capacity compared to shallow networks. Therefore, deep learning techniques can be understood as a large-scale optimization task: similar to an easy problem in terms of formulation but complex due to its size. Moreover, their capacity to adapt directly to the data without any prior assumptions provides significant advantages when dealing with little information about the time series \cite{Lara:2020}. With the increasing availability of data, more sophisticated deep learning architectures have been proposed with substantial improvements in forecasting performances \cite{Gamboa:2017}. However, there is the need more than ever for works that provide a comprehensive analysis of the TSF literature, in order to better understand the scientific advances in the field.

In this work, a thorough review of existing deep learning techniques for TSF is provided. Existing reviews have either focused just on a specific type of deep learning architecture \cite{Hewamalage:2019} or on a particular data scenario \cite{Sezer:2020}. Therefore, in this study, we aim to fill this gap by providing a more complete analysis of successful applications of DL for TSF. The revision of the literature includes studies from different TSF domains considering all the most popular DL architectures (multi-layer perceptron, recurrent, and convolutional). Furthermore, we also provide a thorough experimental comparison between these architectures. We study the performance of seven types of DL models: multi-layer perceptron, Elman recurrent, long short-term memory, echo state, gated recurrent unit, convolutional, and temporal convolutional networks. For evaluating these models, we have used 12 publicly available datasets from different fields such as finance, energy, traffic, or tourism. We compare these models in terms of accuracy and efficiency, analyzing the distribution of results obtained with different hyperparameter configurations. A total of 6432 models have been tested for each dataset, covering a large range of possible architectures and training parameters.

Since novel DL approaches in the literature are often compared to classical models instead of other DL techniques, this experimental study aims to provide a general and reliable benchmark that can be used for comparison in future studies. To the best of our knowledge, this work is the first to assess the performance of all the most relevant types of deep neural networks over a large number of TSF problems of different domains. Our objective in this work is to evaluate standard DL networks that can be directly applicable to general forecasting tasks, without considering refined architectures designed for a specific problem. The proposed architectures used for comparison are domain-independent, intending to provide general guidelines for researchers about how to approach forecasting problems. 

In summary, the main contributions of this paper are the following:
\begin{itemize}
    
\item An updated exhaustive review on the most relevant DL techniques for TSF according to recent studies.
\item A comparative analysis that evaluates the performance of several DL architectures on a large number of datasets of different nature.
\item An open-source deep learning framework for TSF that implements the proposed models.
\end{itemize}

The rest of the paper is structured as follows:
Section \ref{review} provides a comprehensive review of the existing literature on deep learning for TSF; in Section \ref{materials-methods}, the materials used and the methods proposed for the experimental study are described; Section \ref{results} reports and discusses the results obtained; Section \ref{conclusions} presents the conclusions and potential future work.

\section{Deep learning architectures for time series forecasting}
\label{review}
A time series is a sequence of observations in chronological order that are recorded over fixed intervals of time. The forecasting problem refers to fitting a model to predict future values of the series considering the past values (which is known as lag). Let $X = \{x_1,x_2,...,x_T\}$ be the historical data of a time series and $H$ the desired forecasting horizon, the task is to predict the next values of the series $\{x_{T+1}, ...,x_{T+H}\}$. Being $\hat{X} = \{\hat{x}_1,\hat{x}_2,...,\hat{x}_T\}$ the vector of predicted values, the goal is to minimize the prediction error as follows:.

\begin{equation}
    \label{eq:error}
    {E} = \sum_{i=1}^{h=H} \lvert x_{T+i} - \hat{x}_{T+i} \rvert
\end{equation}

Time series can be divided into univariate or multivariate depending on the number of variables at each timestep. In this paper, we deal only with univariate time series analysis, with a single observation recorded sequentially over time. Over the last decades, artificial intelligence (AI) techniques have increased their popularity for approaching TSF problems, with traditional statistical methods being regarded as baselines for performance comparison in novel studies.

Before the rise of data mining techniques, the traditional methods used for TSF were mainly based on statistical models, such as exponential smoothing (ETS) \cite{Hyndman:2008} and Box-Jenkins methods like ARIMA \cite{Box:1994}. These models rely on building linear functions from recent past observations to provide future predictions, and have been extensively used for forecasting tasks over the last decades \cite{Zhang:2003}. However, these statistical methods often fail when they are applied directly, without considering the theoretical requirements of stationarity and ergodicity as well as the preprocessing required.
For instance, in an ARIMA model, the time series has to be transformed into stationary (without trends or seasonality) through several differencing transformations. Another issue is that the applicability of these models is limited to situations when sufficient historical data, with an explainable structure, is available. When the data availability for a single series is scarce, these models often fail to extract effectively the underlying features and patterns. 
Furthermore, since these methods create a model for each individual time series, it is not possible to share the learning across similar instances. Therefore, it is not feasible to use these techniques for dealing with massive amounts of data as it would be computationally prohibitive. Hence, artificial intelligence (AI) techniques, that allow building global models to forecast on multiple related series, began earning popularity. Meanwhile, linear methods started to be regarded as baselines for performance comparison in novel studies.

Ref. \refcite{Martinez-Alvarez:2015} provides a survey on early data mining techniques for TSF and its comparison with classical approaches. Among all proposed data mining methods in the literature, the models that have attracted more attention have been those based on artificial neural networks (ANNs). They have shown better performance than statistical methods in many situations, especially due to their capacity and flexibility to map non-linear relationships from data given their deep structure \cite{Raza:2015}. Another advantage of ANNs is that they can extract temporal patterns automatically without any theoretical assumptions on the data distribution, reducing preprocessing efforts. Furthermore, the capacity of generalization of ANNs allows exploiting cross-series information. 
When using ANNs, a single global model that learns from multiple related time series can be built, which can greatly improve the forecasting performance.

However, researchers have struggled to develop optimal network topologies and learning algorithms, due to the infinite possibilities of architecture configurations that ANNs allow. Starting from very basic Multi-Layer Perceptron (MLP) proposals, a large number of studies have proposed increasingly sophisticated architectures, such as recurrent or convolutional networks, that have enhanced the performance for TSF. In this work, we review the most relevant types of DL networks according to the existing literature, which can be divided into three categories:

\begin{itemize}
    \item[$\bullet$] Fully connected neural networks.
    \begin{itemize}
        \item[--] MLP: Multi-layer perceptron.
    \end{itemize}
    \item[$\bullet$] Recurrent neural networks.
    \begin{itemize}
        \item[--] ERNN: Elman recurrent neural network.
        \item[--] LSTM: Long short-term memory network.
        \item[--] ESN: Echo state network.
        \item[--] GRU: Gated recurrent units network.
    \end{itemize}
    \item[$\bullet$] Convolutional networks.
    \begin{itemize}
        \item[--] CNN: Convolutional neural network.
        \item[--] TCN: Temporal convolutional network.
    \end{itemize}
\end{itemize}

The architectural variants and the fields in which these DL networks have been successfully applied will be discussed in the following sections.

\newcolumntype{P}[1]{>{\centering\arraybackslash}p{#1}}
\newcolumntype{M}[1]{>{\centering\arraybackslash}m{#1}}

\begin{table*}[b]
\tbl{Relevant studies on time series forecasting using MLP networks.
\label{tab:mlp-review}}{
\resizebox{\textwidth}{!}{%
\setlength{\extrarowheight}{.3em}
\begin{tabular}{ccP{2.5cm}P{3.5cm}P{3cm}p{7.5cm}}
\hline
\textbf{Ref.} & \multicolumn{1}{l}{\textbf{Year}} & \textbf{Technique} & \textbf{Outperforms} & \textbf{Application} & \textbf{Description/findings} \\\hline
\refcite{Zhang:2003} & 2003 & Hybrid ARIMA-MLP & ARIMA and MLP individually & Exchange rates and environmental data & The ARIMA component models the linear correlation structures while MLP works on the nonlinear part. \\
\refcite{Palmer:2006} & 2006 & MLP & Statistical methods & Tourism expenditure & Pre-processing steps such as detrending and deseasonalization were essential. MLPs performed better as the forecast horizon increases. \\
\refcite{Zhang:2007} & 2007 & 48 MLPs with different inputs & No comparison  & Quarterly time series M3 competition &
Data preparation was more important than optimizing the number of nodes. Carefully selected the input variables and designed simple models.  \\
\refcite{Hamzacebi:2009} & 2009 & MLP & ARIMA & Six datasets of different domains & Comparison between iterative and direct forecasting methods. Best performance with direct approach. \\
\refcite{Yan:2012} & 2012 & MLP & Competition winner & NN3 competition & Automatic scheme based on generalized regression neural networks (GRNNs). \\
\refcite{Claveria:2014} & 2014 & MLP & ARIMA & Tourism data & ARIMA models were better for short-horizon prediction while ANNs were better for longer forecasts. \\
\refcite{Jourentzes:2014} & 2014 & Ensemble MLP & Exponential smoothing and naive forecast & Retail sales  & Evaluates different ensemble operators (mean, median, and mode). The mode operator was the best. \\
\refcite{Rahman:2016} & 2016 & Ensemble MLP & Statistical methods & NN3 and NN5 competitions &  Two-layers ensemble. The first layer finds an appropriate lag, and the second layer employs the obtained lag for forecasting.\\
\refcite{Torres:2018} & 2018 & Parallelized MLPs & Linear Regression, Gradient-Boosted Trees, Random Forest & Electricity demand & Split
the problem into several forecasting subproblems to predict many samples simultaneously. \\ \hline
\end{tabular}%
}}
\end{table*}

\subsection{Multi-Layer Perceptron}
The concept of ANNs was first introduced in Ref. \refcite{McCulloch:1943}, and was inspired in the functioning of the brain with neurons acting in parallel for processing data. Multi-Layer Perceptron (MLP) is the most basic type of feed-forward artificial neural network. Their architecture is composed of a three-block structure: an input layer, hidden layers, and an output layer. There can be one or multiple hidden layers, which determines the depth of the network. It is the increase of the depth, the inclusion of more hidden layers, which makes an MLP network a deep learning model. Each layer contains a defined set of neurons that have connections associated with trainable parameters. The neural network learning algorithm updates the weights of these connections in order to map the input/output relationship. MLP networks only have forward connections between neurons, unlike other architectures that have feedback loops. The most relevant studies using MLPs are presented below.


In the early 90s, researchers started to pose ANNs as a promising alternative compared to traditional statistical models. These initial studies proposed the use of MLP networks with one or a few hidden layers. 
At that time, due to the lack of systematic methodologies to build ANNs and their difficult interpretation, researchers were still skeptical about their applicability to TSF.

At the end of the decade, several works reviewed the existing literature with the aim of clarifying the strengths and shortcomings of ANNs \cite{Hippert:2001}. 
These reviews agreed about the potential of ANNs for forecasting due to their unique characteristics as universal function approximators and their flexibility to adapt to data without prior assumptions. However, they all claimed that more rigorous validation procedures were needed to conclude generally that ANNs improve the performance of classical alternatives. Furthermore, another general reasoning was that determining the optimal structure and hyperparameters of the ANNs was a difficult process that had a key influence on performance. From that time on, many references proposing ANNs as a powerful tool for forecasting can be found in the literature.

Table \ref{tab:mlp-review} presents a collection of studies using MLPs for different time series forecasting problems. Most studies have focused on the design of the network, concluding that the past-history parameter has a greater influence on the accuracy than the number of hidden layers. Furthermore, these works proved the importance of the preprocessing steps, since simple models with carefully selected input variables tend to achieve better performance. However, the most recent studies propose the use of ensembles of MLP networks to achieve higher accuracy.


\begin{table*}[!b]
\tbl{Relevant studies on time series forecasting using ERNNs.
\label{tab:ernn-review}}{
\resizebox{\textwidth}{!}{%
\setlength{\extrarowheight}{.3em}
\begin{tabular}{ccP{1.5cm}P{3.5cm}P{3cm}p{7.5cm}}
\hline
\textbf{Ref.} & \multicolumn{1}{l}{\textbf{Year}} & \textbf{Technique} & \textbf{Outperforms} & \textbf{Application} & \textbf{Description/findings} \\ \hline
\refcite{Sfetsos:2000} & 2000 & ERNN & MLP, RBF, ARIMA & Solar radiation & Despite the good results, ERNNs were unable to model the discontinuities of time series. ERNNs had slower convergence rate than non-recurrent models. \\
\refcite{CHANDRA:2012} & 2012 & ERNN & RBFNN, ARMA-ANN, NARX & Chaotic time series & Compares different training algorithms, concluding that evolutionary algorithms take significantly more time to converge than gradient-based methods. \\
\refcite{Mohammadi:2018} & 2018 & ERNN & ARIMA, SVR, BPNN, RBFNN & Electricity data & Improved performance with pre-processing steps such as signal decomposition and  feature selection. \\
\refcite{Ruiz:2018} & 2018 & ERNN & NAR, NARX & Energy consumption & Reported an important improvement with respect to nonlinear autoregressive networks. \\ \hline
\end{tabular}%
}}
\end{table*}

Although feed-forward neural networks such as MLP have been applied effectively in many circumstances, they are unable to capture the temporal order of a time series, since they treat each input independently. Ignoring the temporal order of input windows restrains performance in TSF, especially when dealing with instances of different lengths that change dynamically. Therefore, more specialized models, such as recurrent neural networks (RNNs) or convolutional neural networks (CNNs), started to raise interest. With these networks, the temporal problem is transformed into a spatial architecture that can encode the time dimension, thus capturing more effectively the underlying dynamical patterns of time series \cite{Schafer:2006}.

\subsection{Recurrent Neural Networks}

Recurrent Neural Networks (RNN) were introduced as a variant of ANN for time-dependent data. 
While MLPs ignores the time relationships within the input data, RNNs connects each time step with the previous ones to model the temporal dependency of the data, providing RNN native support for sequence data \cite{Elman:1990}.
The network sees one observation at a time and can learn information about the previous observations and how relevant the observation is to forecasting.
Through this process, the network not only learns patterns between input and output but also learns internal patterns between observations of the sequence.
This characteristic makes RNNs ones of the most common neural networks used for time-series data. They have been successfully implemented for forecasting applications in different fields such as stock market price forecasting \cite{Kim:2018}, 
wind speed forecasting \cite{Yu:2018}, or 
solar radiation forecasting \cite{Wang:2019}. 
Furthermore, RNNs have achieved top results at forecasting competitions, like the recent M4 competition \cite{M4:2018}. 

In the late 80s, several studies worked on different approaches to provide memory to a neural network \cite{Doya:1989}. This memory would help the model to learn from time series data. One of the most promising proposals was the Jordan RNN \cite{Jordan:1986}. It was the main inspiration to create the Elman RNN, which is known as the base of modern RNN \cite{Elman:1990}.

\subsubsection{Elman Recurrent Neural Networks}

The ERNN aimed to tackle the problem of dealing with time patterns in data. ERNN changed the feed-forward hidden layer to a recurrent layer, which connects the output of the hidden layer to the input of the hidden layer for the next time step. This connection allows the network to learn a compact representation of the input sequence. In order to implement the optimization algorithm, the networks are usually unfolded and the backpropagation method is modified, resulting in the Truncated Backpropagation Through Time (TBPTT) algorithm. Table \ref{tab:ernn-review}  presents some studies where ERNNs have been successfully applied for TSF in different fields. In general, the studies demonstrate the improvement of recurrent networks compared to MLPs and linear models such as ARIMA.






Despite the success of Elman's approximation, the application of TBTT faces two main problems: the weights may start to oscillate (exploding gradient problem), or an excessive computational time to learn long-term patterns (vanishing gradient problem) \cite{Bengio:1994}

\begin{table*}[b]
\tbl{Relevant studies on time series forecasting using LSTM networks.
\label{tab:lstm-review}}{
\resizebox{\textwidth}{!}{%
\setlength{\extrarowheight}{.3em}
\begin{tabular}{ccP{3cm}P{3.5cm}P{3cm}p{7.5cm}}
\hline
\textbf{Ref.} & \multicolumn{1}{l}{\textbf{Year}} & \textbf{Technique} & \textbf{Outperforms} & \textbf{Application} & \textbf{Description/findings} \\ \hline
\refcite{Ma:2015} & 2015 & LSTM & MLP, NARX, SVM & Traffic speed & LSTM proved effective for short-term prediction without prior information of time lag. \\
\refcite{Tian:2015} & 2015 & LSTM & MLP, Autoencoders & Traffic flow  & LSTM determines the optimal time lags dynamically, showing higher generalization ability. \\
\refcite{Fischer:2018} & 2018 & LSTM & Logistic regression, MLP, RF  & S\&P 500 index  & Disentangle the LSTM black-box to find common patterns of stocks in noisy financial data. \\
\refcite{Bouktif:2018} & 2018 & LSTM & Linear regression, kNN, RF, MLP & Electric load & Feature selection and genetic algorithm to find optimal time lags and number of layers for LSTM model. \\
\refcite{Sagheer:2019} & 2019 & LSTM & ARIMA, ERNN, GRU & Petroleum production & Deep LSTM using genetic algorithms to optimally configure the architecture. \\
\refcite{tan:2019} & 2019 & LSTM + Attention & LSTM, MLP & Solar generation & Temporal attention mechanism to improve performance over standard LSTM, also using partial autocorrelation to determine the input lag. \\ 
\refcite{Smyl:2020} & 2020 & Hybrid ETS-LSTM &  Competition winner & M4 competition & ETS captures the main components such as seasonality, while the LSTM networks allow non-linear trends and cross-learning from multiple related series. \\ 
\refcite{Bandara:2020} & 2020 & Clustering + LSTM & LSTM, ARIMA, ETS & CIF2016 and NN5 competitions & Building a model for each cluster of related time series can improve the forecast accuracy, while also reducing training time. \\ \hline
\end{tabular}%
}}
\end{table*}

\begin{table*}[b]
\tbl{Relevant studies on time series forecasting using ESNs.
\label{tab:esn-review}}{
\resizebox{\textwidth}{!}{%
\setlength{\extrarowheight}{.3em}
\begin{tabular}{ccP{2.5cm}P{3.8cm}P{3cm}p{7.5cm}}
\hline
\textbf{Ref.} & \multicolumn{1}{l}{\textbf{Year}} & \textbf{Technique} & \textbf{Outperforms} & \textbf{Application} & \textbf{Description/findings} \\ \hline
\refcite{Lin:2009} & 2009 & ESN & MLP, RBFNN & Stock market price  & Applying PCA to filter noise improved the ESN performance over some indexes. \\
\refcite{Rodan:2011} & 2011 & ESN & Other types of reservoirs & 7 time-series from different fields & An ESN with a simple cycle reservoir topology achieved high performance for TSF problems. \\
\refcite{Li:2012} & 2012 & ESN with Laplace function & Support Vector, Gaussian, and Bayesian ESN & Simulated datasets & Applied ESN in a Bayesian learning framework. The Laplace function proved to be more robust to outliers than the Gaussian distribution. \\
\refcite{Deihimi:2012} & 2012 & ESN & No comparison  & Electricity load forecasting & The ESN proved its capacity to learn complex dynamics of electric load, obtaining high accuracy results without additional inputs. \\
\refcite{Liu:2015} & 2015 & GA-optimized ESN & ARIMA, Generalized Regression NN & Wind speed  & ESN to predict multiple wind speeds simultaneously. PCA and spectral clustering to preprocess data and genetic algorithm to search optimal parameters. \\
\refcite{Hu:2020} & 2020 & Bagged ESN & BPNN, RNN, LSTM & Energy consumption & Combines ESN, bagging, and differential evolution algorithm to reduce forecasting error and improve generalization capacity. \\ \hline
\end{tabular}%
}}
\end{table*}

\subsubsection{Long Short-Term Memory Networks}\label{lstm}

In 1997, Long Short-Term Memory (LSTM) networks \cite{LSTM:1997} were introduced as a solution for ERNN's problems. 
LSTMs are able to model temporal dependencies in larger horizons without forgetting the short-term patterns.  
LSTM networks differ from ERNN in the hidden layer, also known as LSTM memory cell. LSTM cells use a multiplicative input gate to control the memory units, preventing them from being modified by irrelevant perturbations. Similarly, a multiplicative output protects other cells from perturbations stored in the current memory. Later, another work added a forget gate to the LSTM memory cell \cite{Gers:2000}.
This gate allows LSTM to learn to reset memory contents when they become irrelevant.

A list of relevant studies that address TSF problems with LSTM networks can be found in Table \ref{tab:lstm-review}. Overall, these studies prove the advantages of LSTM over traditional MLP and ERNN for extracting meaningful information from time series. Furthermore, some recent studies proposed more innovative solutions such as hybrid models, genetic algorithms to optimize the network architecture, or adding a clustering step to train LSTM networks on multiple related time series.





\begin{table*}[!b]
\tbl{Relevant studies on time series forecasting using GRU networks.
\label{tab:gru-review}}{
\resizebox{\textwidth}{!}{%
\setlength{\extrarowheight}{.3em}
\begin{tabular}{ccP{3cm}P{3cm}P{3cm}p{7.5cm}}
\hline
\textbf{Ref.} & \multicolumn{1}{l}{\textbf{Year}} & \textbf{Technique} & \textbf{Outperforms} & \textbf{Application} & \textbf{Description/findings} \\ \hline
\refcite{Chung:2014} & \multicolumn{1}{l}{2014} & GRU & ERNN &  Music and speech signal modeling & The results proved the advantages of GRU and LSTM networks over ERNN but did not show a significant difference between both gated units. \\
\refcite{Kuan:2017} & \multicolumn{1}{l}{2017} & GRU & LSTM, GRU & Electricity load &
Proposed scaled exponential linear units to overcome vanishing gradients, showing a significant improvement over LSTM and standard GRU models. \\
\refcite{WangLiao:2018} & \multicolumn{1}{l}{2018} & GRU & LSTM, SVM, ARIMA & Photovoltaic power & Pearson coefficient is used to extract the main features and K-means to cluster similar groups that train the GRU model. \\
\refcite{Ugurlu:2018} & 2018 & GRU & MLP, CNN, LSTM & Electricity price & GRU performed better and trained faster than LSTM networks. Using a larger number of lagged values improved the results. \\ \hline
\end{tabular}%
}}
\end{table*}
\subsubsection{Echo State Networks}
Echo State Networks (ESNs) were introduced by Ref. \refcite{ESN:2001}. They are based on Reservoir Computing (RC), which simplifies the training procedure of traditional RNNS. Previous RNNs, such as ERNN or LSTM, have to find the best values for all neurons of the network. In contrast, the ESN tunes just the weights from the output neurons, which makes the training problem a simple linear regression task \cite{IsmailFawaz:2018}.



An ESN is a neural network with a random RNN called the \textit{reservoir} as the hidden layer. The reservoir has usually a high number of neurons and sparse interconnectivity (around 1\%) between the hidden neurons. These characteristics make the reservoir a set of subsystems that work as echo functions, being able to reproduce specific temporal patterns \cite{Jaeger:2004}. 

In the literature, many ESN applications for TSF can be found. Especially, ESN networks have proved to outperform MLPs and statistical methods when modeling chaotic time series data. Furthermore, it is worth mentioning that the non-trainable reservoir neurons make this network a very time-efficient model compared to other RNNs. Table \ref{tab:esn-review} presents several studies using ESNs and their most interesting findings.

\subsubsection{Gated Recurrent Units}
Gated Recurrent Units (GRU) were introduced by Ref. \refcite{GRU:2014} as another solution to the exploding gradient and vanishing gradient problem of ERNNs. However, it can also be seen as a simplification of LSTM units. In a GRU unit, the forget and input are combined into a single update gate. This modification reduces the trainable parameters providing GRU networks with a better performance in terms of computational time while achieving similar results \cite{Ravanelli:2018}.

The GRU cell uses an update gate and a reset gate that will learn to decide which information should be kept without vanishing it through time and which information is irrelevant for the problem. The update gate is in charge of deciding how much of the past information should be passed along to the future, while the reset gate decides how much of the past information to forget. 

Table \ref{tab:gru-review} presents several studies that use GRU networks for TSF problems. These works often propose modifications to the standard GRU model in order to improve performance over other recurrent networks. However, the number of existing studies proposing GRU models is much lower than those using LSTM networks.






\subsection{Convolutional Neural Networks}

\begin{table*}[tb]
\tbl{Relevant studies on time series forecasting using CNNs.
\label{tab:cnn-review}}{
\resizebox{\textwidth}{!}{%
\setlength{\extrarowheight}{.3em}
\begin{tabular}{ccP{3cm}P{3cm}P{3cm}p{7.5cm}}
\hline
\textbf{Ref.} & \multicolumn{1}{l}{\textbf{Year}} & \textbf{Technique} & \textbf{Outperforms} & \textbf{Application} & \textbf{Description/findings} \\ \hline
\refcite{Tsantekidis:2017} & 2017 & CNN & SVM, MLP & Stock price & The data was preprocessed using temporally-aware normalization. The model obtained better performance in short-term predictions. \\
\refcite{Kuo:2018} & 2018 & CNN & LSTM, MLP & Energy load & The model used convolution and pooling to predict the load for the next three days. It proved useful for reducing expenses in future smart grids. \\
\refcite{Koprinska:2018} & 2018 & CNN & LSTM & Solar power and electricity & The CNN was significantly faster than the recurrent approach with similar accuracy, hence more suitable for practical applications. \\
\refcite{Tian:2018} & 2018 & Hybrid CNN-LSTM & RNN and LSTM individually & Electricity & The LSTM is able to extract the long-term dependencies and the CNN captures local trend patterns. \\
\refcite{Liu:2018} & 2018 & Ensemble CNN-LSTM & ARIMA, ERNN, RBF & Wind speed & The data is decomposed using the wavelet packet, with a CNN predicting the high-frequency data and a CNN-LSTM for the low-frequency data. \\
\refcite{Cai:2019} & 2019 & CNN & RNN, ARIMAX & Building load & The CNN model with a direct approach provided the best results, improving significantly the forecasting accuracy of the seasonal ARIMAX model. \\
\refcite{Shen:2019} & 2019 & Hybrid CNN-LSTM & LSTM, CNN & Financial data &
The LSTM learns features and reduces dimensionality, and the dilated CNN learns different time intervals. \\ \hline
\end{tabular}%
}}
\end{table*}

\begin{table*}[!b]
\tbl{Relevant studies on time series forecasting using TCNs.
\label{tab:tcn-review}}{
\resizebox{\textwidth}{!}{%
\setlength{\extrarowheight}{.3em}
\begin{tabular}{ccP{3cm}P{3cm}P{3cm}p{7.5cm}}
\hline
\textbf{Ref.} & \multicolumn{1}{l}{\textbf{Year}} & \textbf{Technique} & \textbf{Outperforms} & \textbf{Application} & \textbf{Description/findings} \\ \hline
\refcite{Borovykh:2019} & 2019 & TCN & LSTM & Financial data & TCN can effectively learn dependencies in and between series without the need for long historical data. \\
\refcite{Chen:2019} & 2019 & Encoder-decoder TCN & RNN & Retail sales &
Combined with representation learning, TCN can learn complex patterns such as seasonality. TCNs are efficient due to the paralellization of convolutions.
\\
\refcite{Wan:2019} & 2019 & TCN & LSTM, ConvLSTM & Meteorology &  The TCN presented higher efficiency and capacity of generalization, performing better at longer forecasts. \\
\refcite{Lara-Benitez:2020} & 2020 & TCN & LSTM & Energy demand & TCN models showed a greater capacity to deal with longer input sequences. TCNs were less sensitive to the parametrization than LSTM networks. \\ \hline
\end{tabular}%
}}
\end{table*}

CNNs are a family of deep architectures that were originally designed for computer vision tasks. They are considered state-of-the-art for many classification tasks such as object recognition \cite{Kang:2018}, speech recognition \cite{Hinton:2012} and natural language processing \cite{Kim:2014}. CNNs can automatically extract features from high dimensional raw data with a grid topology, such as the pixels of an image, without the need of any feature engineering. The model learns to extract meaningful features from the raw data using the convolutional operation, which is a sliding filter that creates feature maps and aims to capture repeated patterns at different regions of the data. This feature extracting process provides CNNs with an important characteristic so-called distortion invariance, which means that the features are extracted regardless of where they are in the data. These characteristics make CNNs suitable for dealing with one-dimensional data such as time series. Sequence data can be seen as a one-dimensional image from where the convolutional operation can extract features.

A CNN architecture is usually composed of convolution layers, pooling layers (to reduce the spatial dimension of feature maps), and fully connected layers (to combine local features into global features). These networks are based on three principles: local connectivity, shared weights, and translation equivariance. Unlike standard MLP networks, each node is connected only to a region of the input, which is known as the receptive field. Moreover, the neurons in the same layers share the same weight matrix for the convolution, which is a filter with a defined kernel size. These special properties allow CNNs to have a much smaller number of trainable parameters compared to a RNN, hence the learning process is more time efficient \cite{Borovykh:2019}. Another key aspect of the success of CNNs is the possibility of stacking different convolutional layers so that the deep learning model can transform the raw data into an effective representation of the time series at different scales \cite{yang:2015}.


CNN-based models have not been extensively used in the TSF literature since RNNs have been given far more importance. Nevertheless, several works have proposed CNNs as feature extractors alone or together with recurrent blocks to provide forecasts. The most relevant works involving these proposals are detailed in Table \ref{tab:cnn-review}.

\subsubsection{Temporal Convolutional Network}


Recent studies have proposed a more specialized CNN architecture known as Temporal Convolutional Network (TCN). This architecture is inspired by the Wavenet autoregressive model, which was designed for audio generation problems \cite{Oord:2016}. The term TCN was first presented in Ref. \refcite{Bai:2018} to refer to a type of CNN with special characteristics: the convolutions are causal to prevent information loss, and the architecture can process a sequence of any length and map it to an output of the same length. To enable the network to learn the long-term dependencies present in time series, the TCN architecture makes use of dilated causal convolutions. This convolution increases the receptive field of the network (neurons that are convolved with the filter) without losing resolution since pooling operation is not needed \cite{Yu:2016}. 
Furthermore, TCN employs residual connections to allow to increase the depth of the network, so that it can deal effectively with a large history size. In Ref. \refcite{Bai:2018} the authors present an experimental comparison between generics RNNs (LSTM and GRU) and TCNs over several sequence modeling tasks. This work emphasizes the advantages of TCNs such as: their low memory requirements for training due to the shared convolutional filters; long input sequences can be processed with parallel convolutions instead of sequentially as in RNNs; and a more stable training scheme, hence avoiding vanishing gradient problems.

TCNs are acquiring increasing popularity in recent years due to their suitability for dealing with temporal data. Table \ref{tab:tcn-review} presents some studies where TCNs have successfully applied for time series forecasting.

\section{Materials and methods}
In this section, we present the datasets used for the experimental study and the details of the architectures and hyperparameter configurations studied. The source code of the experiments can be found at Ref. \refcite{github-code}. This repository provides a deep learning framework based on TensorFlow for TSF, allowing full reproducibility of the experiments.

\begin{figure*}[t]
    \includegraphics[width=\textwidth]{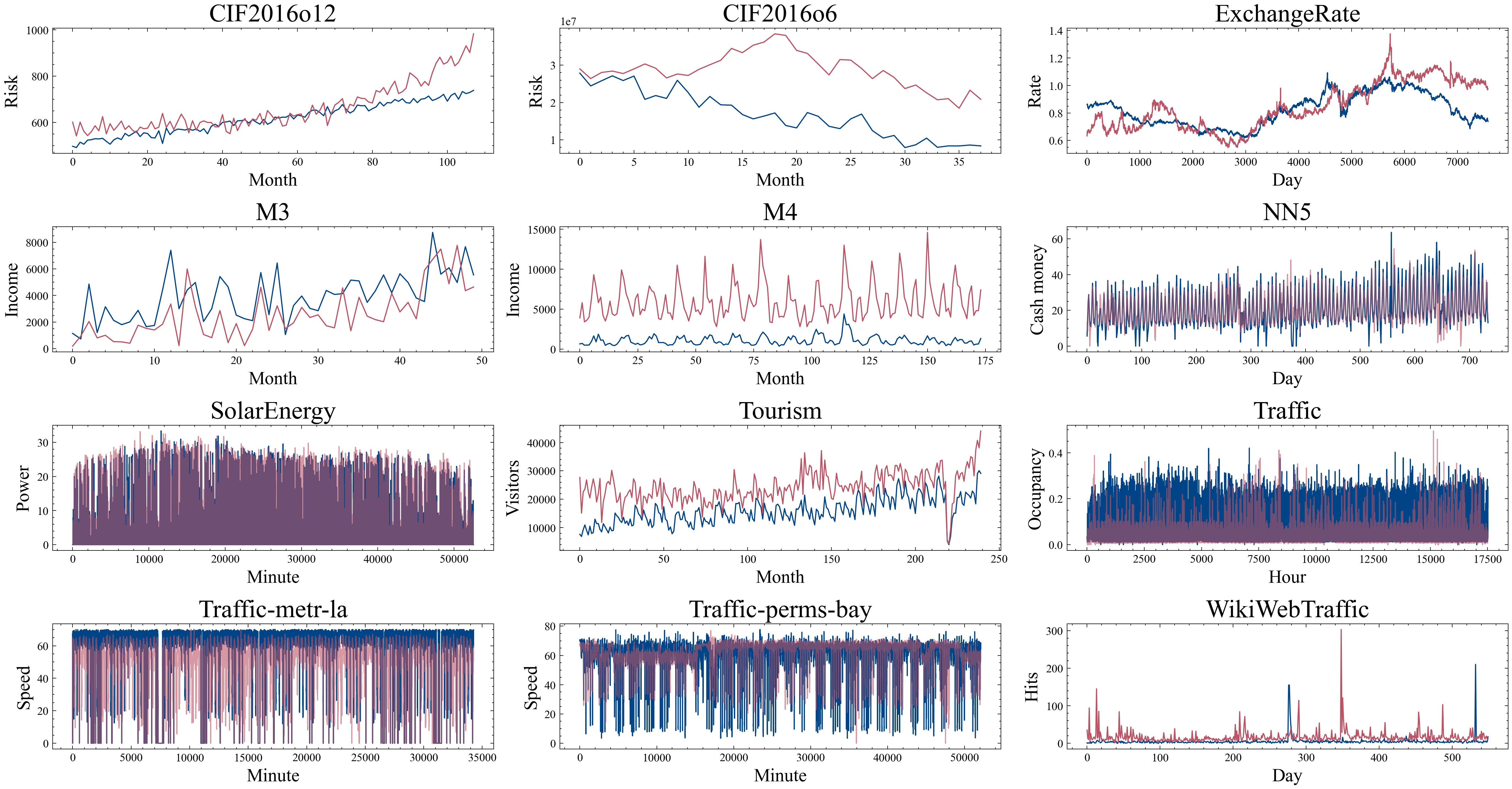}
    \caption{Two examples of time series instances of each data set. The colours differentiate the two instances. The y-axis represents the value that the time series takes for each timestep along the x-axis.}
    \label{fig:example_instances}
\end{figure*}

\begin{table}[H]
\tbl{Datasets used for the experimental study. Columns N, FH, M and m refer to number of time series, forecast horizon, maximum length and minimum length respectively.
\label{table:datasets}}{
\resizebox{\linewidth}{!}{%
\begin{tabular}{@{}cccccc@{}}
\toprule
\textbf{\#} & \textbf{Datasets} & \textbf{N} & \textbf{FH} & \textbf{M} & \textbf{m} \\ \hline
1 & CIF2016o12 & 57 & 12 & 108 & 48 \\
2 & CIF2016o6 & 15 & 6 & 69 & 22 \\
3 & ExchangeRate & 8 & 6 & 7588 & 7588 \\
4 & M3 & 1428 & 18 & 126 & 48 \\
5 & M4 & 48000 & 18 & 2794 & 42 \\
6 & NN5 & 111 & 56 & 735 & 735 \\
7 & SolarEnergy & 137 & 6 & 52560 & 52560 \\
8 & Tourism & 336 & 24 & 309 & 67 \\
9 & Traffic & 862 & 24 & 17544 & 17544 \\
10 & Traffic-metr-la & 207 & 12 & 34272 & 34272 \\
11 & Traffic-perms-bay & 325 & 12 & 52116 & 52116 \\
12 & WikiWebTraffic & 997 & 59 & 550 & 550 \\ \hline
\end{tabular}%
}}
\end{table}

\label{materials-methods}
\subsection{Datasets}
In this study, we have selected 12 publicly available datasets of different nature, each of them with multiple related time series. This collection represents a wide variety of time series forecasting problems, covering different sizes, domains, time-series length, and forecasting horizon.  Table \ref{table:datasets} presents the characteristics of each dataset in detail. In total, the experimental study involves more than 50000 time series among all the selected datasets. Furthermore, Figure \ref{fig:example_instances} illustrates some examples instances of each dataset. As can be seen, the selected datasets present a wide diversity of characteristics in terms of scale and seasonality. Most of these datasets have been used in forecasting competitions and other TSF reviews such as Ref. \refcite{Hewamalage:2019}.

The CIF 2016 competition dataset \cite{CIF:data} contains a total of 72 monthly time series, from which 12 of them have a 6-month forecasting horizon while the remaining 57 series have a 12-month forecasting horizon. Some of these time series are real bank risk analysis indicators while others are artificially generated.

The \textit{ExchangeRate} dataset \cite{ExchangeRate:data} records the daily exchange rates of 8 countries including Australia, British, Canada, Switzerland, China, Japan, New Zealand, and Singapore from 1990 to 2016 (a total of 7588 days for each country). The goal of this dataset is to predict values for the next 6 days.

Both \textit{M3} and \textit{M4} competitions \cite{M3:data:paper, M4:data:paper} datasets include time series of different domains and observation frequencies. However, due to computational time constraints, we have limited the experimentation to the monthly time series as they are longer than the yearly, quarterly, weekly and daily time series. Both competitions ask for an 18 months prediction and contain time series of different categories such as industry, finance, or demography. The number of instances belonging to each category are evenly distributed according to their presence in the real world, leading novel studies to representative conclusions. Considering only the monthly time series, the M4 dataset has 48000 time series, while M3 is considerably smaller with 1428 time series.

The \textit{NN5} competition dataset \cite{NN5:2008} is formed by 111 time series with a length of 735 values. This dataset represents 2 years of daily cash withdrawals at automatic teller machines (ATMs) from England. The competition established a forecasting horizon of 56 days ahead. 

The \textit{Tourism} dataset \cite{Tourism:data} is composed of 336 monthly time series of different length, requiring a 24 months prediction. The data was provided by tourism bodies from Australia, Hong Kong, and New Zealand, representing total tourism numbers at a country level.

The \textit{SolarEnergy} dataset \cite{SolarEnergy:data} contains the solar power production records in the year of 2006, which is sampled every 10 minutes from 137 solar photovoltaic power plants in Alabama State. The forecasting horizon has been established to one hour, which comprises 6 observations. 

\textit{Traffic-metr-la} and \textit{Traffic-perms-bay} are two public traffic network datasets \cite{Traffic-la-bay:data}. Traffic-metr-la contains traffic speed information of 207  sensors on the highways of  Los  Angeles for four months. Similarly, Traffic-perms-bay records six months of statistics on traffic speed from 325 sensors in the Bay area. For both datasets, the readings from the sensors are aggregated into five-minute windows. 

The \textit{Traffic} dataset is a collection of hourly time series from the California Department of Transportation collected during 2015 and 2016. The data describes the road occupancy rates (between 0 and 1) measured by different sensors on San Francisco Bay area free-ways.  

The \textit{WikiWebTraffic} dataset belongs to a Kaggle competition \cite{WikiWebTraffic:data} with the goal of predicting future web traffic of a given set of Wikipedia pages. The traffic is measured in the daily number of hits and the forecasting horizon is 59 days.

\subsection{Experimental setup} 
This subsection presents the comparative study carried out to evaluate the performance of seven state-of-the-art deep-learning architectures for time series forecasting. We explain the architectures and hyperparameter configurations that we have explored, together with the details of the evaluation procedure.
The experimental study is based on a statistical analysis with the results obtained with 6432 different architecture configurations over 12 datasets, resulting in more than 38000 tested models.

\subsubsection{Deep learning architectures} 

Seven different types of deep learning models for TSF are compared in this experimental study: multi-layer perceptron, Elman recurrent, long short-term memory, echo state, gated recurrent unit, convolutional, and temporal convolutional networks. For all these models, the number of hyperparameters that have to be configured is high compared to traditional machine-learning techniques. Therefore, the proper tuning of these parameters is a complex task that requires considerable expertise and is usually driven by intuition. In this work, we have performed an exhaustive grid search on the configuration of each type of architecture in order to find the most suitable values. Table \ref{tab:models-params} presents the search carried out over the main parameters of the deep learning models such as the number of layers or units. The grid of possibilities has been decided based on typical values found in the literature. For instance, it is a common practice to select powers of two for the number of neurons, units, or filters. We have tried to establish a fair comparison between architectures, maintaining the values consistent across the models as long as possible. For example, in all cases, we explore single-layer models up to deeper networks with four stacked layers.

\begin{table}[H]
\tbl{Parameter grid of the architecture configurations for the seven types of deep learning models.
\label{tab:models-params}}{
\resizebox{\linewidth}{!}{%
\begin{tabular}{ccc}
\hline
\textbf{Models} & \textbf{Parameters} & \textbf{Values} \\ \hline
\textbf{MLP} & Hidden Layers & \begin{tabular}[c]{@{}c@{}}{[}8{]}, {[}8, 16{]}, {[}16, 8{]},  {[}8, 16, 32{]},\\ {[}32, 16, 8{]}, {[}8, 16, 32, 16, 8{]}, \\ {[}32{]}, {[}32, 64{]}, {[}64, 32{]}, \\ {[}32, 64, 128{]}, {[}128, 64, 32{]},\\ {[}32, 64, 128, 64, 32{]}\end{tabular} \\
\rowcolor[HTML]{F3F3F3} 
\cellcolor[HTML]{F3F3F3} & Layers & 1, 2, 4 \\
\rowcolor[HTML]{F3F3F3} 
\cellcolor[HTML]{F3F3F3} & Units & 32, 64, 128 \\
\rowcolor[HTML]{F3F3F3} 
\multirow{-3}{*}{\cellcolor[HTML]{F3F3F3}\textbf{ERNN}} & Return sequence & True, False \\
 & Layers & 1, 2, 4 \\
 & Units & 32, 64, 128 \\
\multirow{-3}{*}{\textbf{LSTM}} & Return sequence & True, False \\
\rowcolor[HTML]{F3F3F3} 
\cellcolor[HTML]{F3F3F3} & Layers & 1, 2, 4 \\
\rowcolor[HTML]{F3F3F3} 
\cellcolor[HTML]{F3F3F3} & Units & 32, 64, 128 \\
\rowcolor[HTML]{F3F3F3} 
\multirow{-3}{*}{\cellcolor[HTML]{F3F3F3}\textbf{GRU}} & Return sequence & True, False \\
 & Layers & 1, 2, 4 \\
 & Units & 32, 64, 128 \\
\multirow{-3}{*}{\textbf{ESN}} & Return sequence & True, False \\
\rowcolor[HTML]{F3F3F3} 
\cellcolor[HTML]{F3F3F3} & Layers & 1, 2, 4 \\
\rowcolor[HTML]{F3F3F3} 
\cellcolor[HTML]{F3F3F3} & Filters & 16, 32, 64 \\
\rowcolor[HTML]{F3F3F3} 
\multirow{-3}{*}{\cellcolor[HTML]{F3F3F3}\textbf{CNN}} & Pool size & 0, 2 \\
 & Layers & 1, 3 \\
 & Filters & 32, 64 \\
 & Dilations & {[}1, 2,  4, 8{]}, {[}1, 2,  4, 8, 16{]} \\
 & Kernel size & 3, 6 \\
\multirow{-5}{*}{\textbf{TCN}} & Return sequence & True, False \\ \hline
\end{tabular}%
}}
\end{table}

Firstly, we experiment with the simplest neural network architecture, the multi-layer perceptron. The MLP models can be used as a baseline for more-complex architectures that obtain a better performance. In particular, we have implemented 12 MLP models, which differ in the number of hidden layers and the number of units in each layer. The selected configurations can be seen in Table \ref{tab:models-params}, where the hidden-layers parameters are defined as a list $[v_1, v_2, ..., v_i, ..., v_n]$. A value $v_i$ from the list represents the number of units in the $i$-th hidden layer. The number of neurons in each layer ranges from 8 to 128, which aims to suit shorter and longer past-history input sequences. Furthermore, the design considers both encoder and decoder-like structures, with more neurons in initial layers and less at the end, and vice versa.  

Concerning recurrent architectures, four different types of models have been implemented in this study: Elman recurrent neural network (ERNN), long-short term memory (LSTM), gated recurrent unit (GRU), and echo state network (ESN). A grid search on the three main parameters is performed, resulting in 18 models for each architecture. 
These parameters are the number of stacked recurrent layers, the number of recurrent units in each layer, and whether the last layer returns the state sequence or just the final state. If return sequence is False, the last layer returns one value for each recurrent unit. In contrast, if return sequence is True, the recurrent layer returns the state of each unit for each timestep, resulting in a matrix of shape $({input\ timesteps} \times {number\ of\ units})$. Finally, as we are working with a multi-step-ahead forecasting problem, the output of the recurrent block is connected to a dense layer with one neuron for each prediction.
Table \ref{tab:models-params} shows all the possible values for each parameter. The number of units ranges from 32 to 128, aiming to explore the convenience of having more or less learning cells in parallel.

Furthermore, 18 convolutional models have been implemented (CNN). These models are formed by stacked convolutional blocks, which are composed of a one-dimensional convolutional layer followed by a max-pooling layer. In Table \ref{tab:models-params}, we present the value search for each parameter. The convolutional blocks are implemented with decreasing kernel sizes, as it is common in the literature. Single-layer models have a kernel of size 3, two-layers models have kernel sizes of 5 and 3, and four-layers models have kernels of size 7-5-3, and 3.
Since the input sequences in the studied datasets are not excessively long (ranging approximately from 20 to 300 timesteps), we have only considered a pooling factor of 2.

Finally, we have also experimented with the temporal convolutional network (TCN) architecture. TCN models are mainly defined by five parameters: number of TCN layers, number of convolutional filters, dilation factors, convolutional kernel size, and whether to return the last output or the full sequence. A grid search on these parameters has been done with the values specified in Table \ref{tab:models-params}, resulting in 32 different TCN architectures. The number of dilations and the kernel sizes have been selected according to the receptive field of the TCN, which follows the formula $({number\ of\ layers} \times {kernel\ size} \times {last\ dilation})$. With the selected grid we cover receptive fields ranging from 24 to 288, which suits the variable length of the input sequences of the studied datasets.

\subsection{Evaluation procedure}

For evaluation purposes, we have divided the datasets into training and test sets. We have followed a fixed origin testing scheme, which is the same splitting procedure as in recent works dealing with datasets containing multiple related time series \cite{Hewamalage:2019}. The test set consists of the last part of each individual time series within the dataset, hence the length is equal to the forecast horizon. Consequently, the remaining part of the time series is used as training data. This division has been applied equally to all datasets, obtaining a total of more than one million timesteps for testing and more than 210 million for training. In this study, we use thousands of time series with a wide variety of forecasting horizons over more than 38000 models. Therefore, this fixed origin validation procedure can lead to representative results and findings.

Once we have split the time series into training and test, we perform a series of preprocessing steps. Firstly, a normalization method is used to scale each time series of training data between 0 and 1, which helps to improve the convergence of deep neural networks. Then, we transform the time series into training instances to be fed to the models. There are several techniques available for this purpose \cite{Bentaieb:2012}: the~recursive strategy, which performs one-step predictions and feeds the result as the last input for the next prediction; the~direct strategy, which builds one model for each time step; and the multi-output approach, which outputs the complete forecasting horizon vector using just one model. In this study, we have selected the Multi-Input Multi-Output (MIMO) strategy. Recent studies show that MIMO outperforms single-output approaches because, unlike the recursive strategy, it does not accumulate the error over the predictions. It is also more efficient in terms of computation time than the direct strategy since it uses one single model for every prediction \cite{Bentaieb:2012}.

\begin{figure}[H]
    \includegraphics[width=1\linewidth]{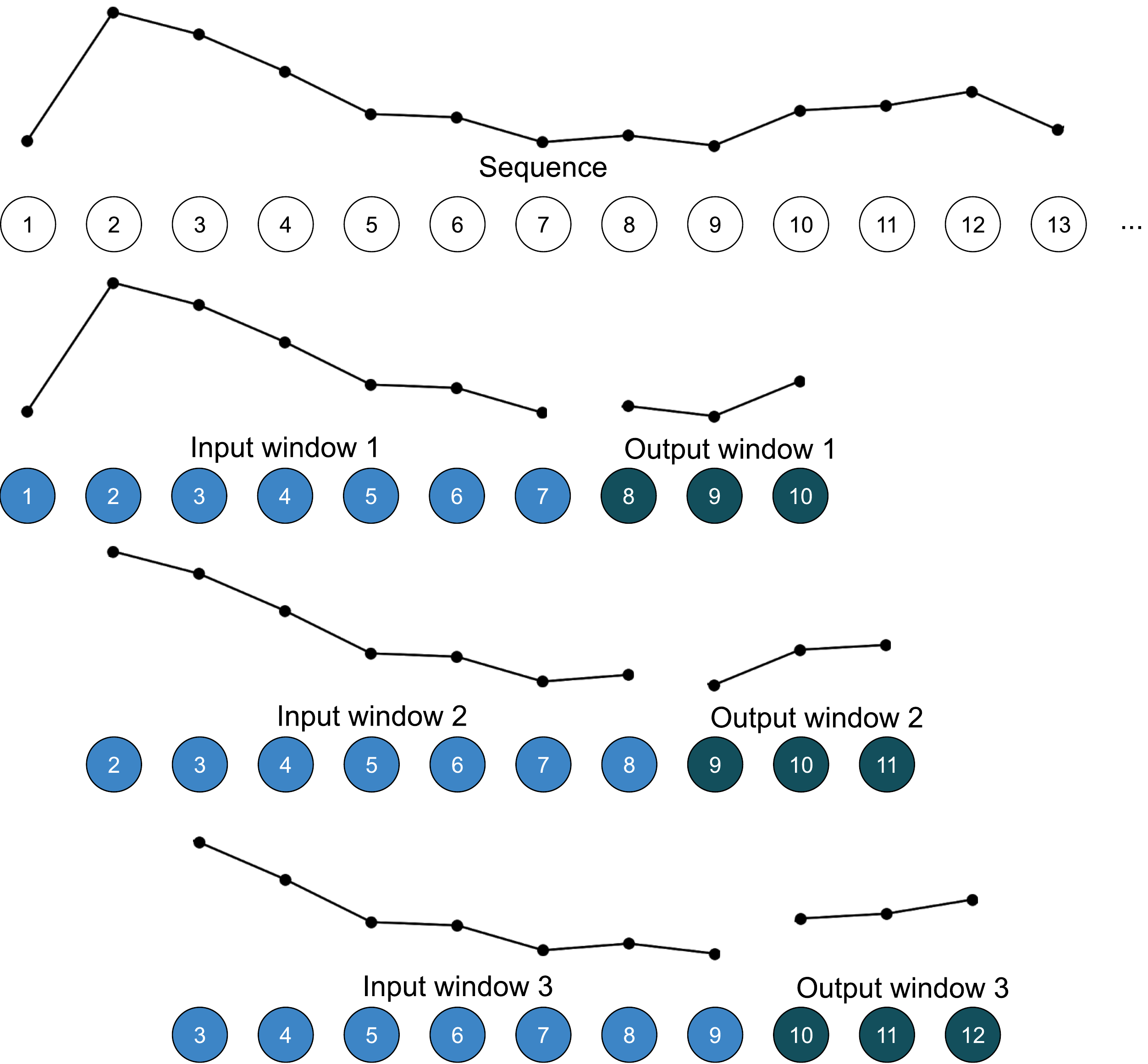}
    \caption{Example of moving window procedure that obtains the input--output instances to train the models. In this example, the models receive an instance with past-history window of length 7 as input and output 3 values as forecasting horizon.}
   \label{fig:MovingWindow}
\end{figure}

Following this approach, we use a moving-window scheme to transform the time series into training instances that can feed the models. This process slides a window of a fixed size, resulting in an input--output instance at each position. The deep learning models receive a window of fixed-length (past history) as input and have a dense layer as output with as many neurons as the forecasting horizon defined by the problem. The length of the input depends on the past-history parameter that should be decided. For the experimentation, we study three different values of past history depending on the forecast horizon (1.25, 2, or 3 times the forecast horizon). Figure \ref{fig:MovingWindow} illustrates an example of this technique with 7 observations as past history and a forecasting horizon of 3 values. 

Along with the past-history parameter, we have also conducted a grid search over the optimal training configuration of the deep learning models. We have experimented with several possibilities for the batch size, the learning rate, and the normalization method to be used. We have selected commonly used values for the batch size (32 and 64) and learning rate (0.001 and 0.01). The Adam optimizer has been chosen, which implements an adaptive stochastic  optimization  method that has been proven to be robust and well-suited for a wide range of machine learning problems\cite{Kingma:2014}. Furthermore, the two most common normalization functions in the literature have been selected to preprocess the data : min-max scaler (Eq. \ref{eq:minmax}) and mean normalization, also known as z-score (Eq. \ref{eq:zscore}).
The complete grid of training parameters can be found in Table \ref{tab:train-params}.

\begin{equation}
    \label{eq:minmax}
    \text{min-max}(x) = \frac{x - \min(x)}{\max(x) - \min(x)}
\end{equation}

\begin{equation}
    \label{eq:zscore}
    \text{z-score}(x) = \frac{x - average(x)}{\max(x) - \min(x)}
\end{equation}

\begin{table}[H]
\tbl{Grid of training parameters for the deep learning models.
\label{tab:train-params}}{
\begin{tabular}{cc}
\hline
\textbf{Parameters} & \textbf{Values} \\ \hline
Past History & (1.25, 2.0, 3.0) $ \times $ Forecast horizon \\
Batch size & 32, 64 \\
No. of epochs & 5 \\
Optimizer & Adam \\
Learning rate & 0.001, 0.01 \\
Normalization & minmax, zscore \\ \hline
\end{tabular}}
\end{table}

\subsubsection{Evaluation metrics}

The performance of the proposed models is evaluated in terms of accuracy and efficiency. We analyze the best results obtained with each type of architecture, as well as the distribution of results obtained with the different hyperparameter configurations. 
For evaluating the predictive performance of all models we use the weighted absolute percentage error (WAPE). This metric is a variation of the mean absolute percentage error (MAPE), which is one of the most widely used measures of forecast accuracy due to its advantages of scale-independency and interpretability \cite{KIM:2016}. WAPE is a more suitable alternative for intermittent and low-volume data. It rescales the error dividing by the mean to make it comparable across time series of varying scales.

WAPE can be defined as follows:
\begin{equation}
    WAPE(y, o) = \frac{MAE(y, o)}{mean(y)} =  \frac{mean(|y - o|)}{mean(y)}, 
\end{equation}
where $y$ and $o$ are two vectors with the real and predicted values, respectively, that have a length equal to the forecasting horizon

Furthermore, developing efficient models is essential in TSF since many applications require real-time responses. Therefore, we also study the average training and inference times of each model.

\begin{figure*}[!b]
    \includegraphics[width=0.999916\textwidth]{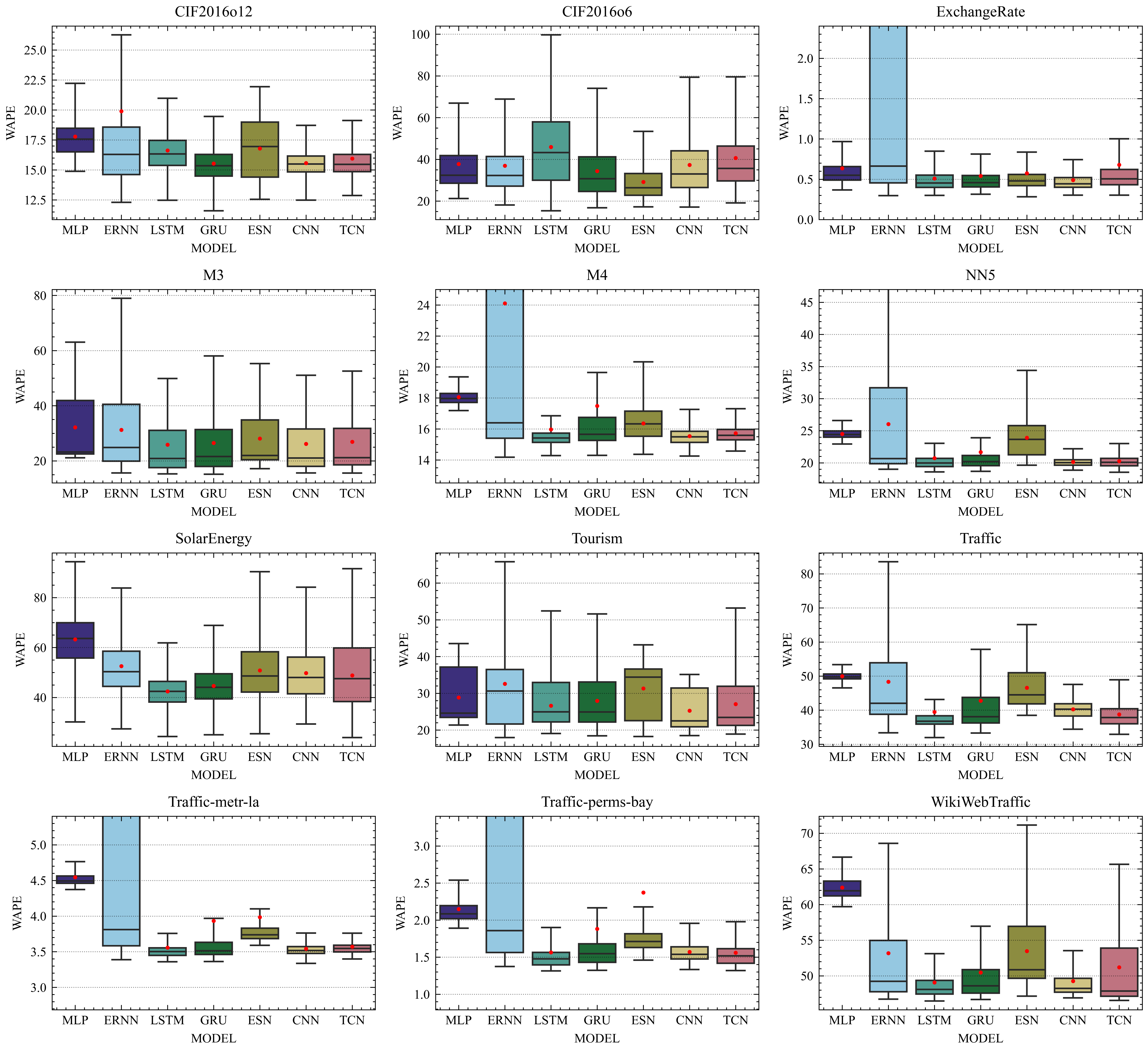}
    \caption{Boxplots showing the distribution of WAPE accuracy results for each type of model over each dataset. The red dot indicates the mean WAPE, the box shows the quartiles of the results and the whiskers extend to show the rest of the distribution.}
    \label{fig:boxplot-wape}
    \vspace*{0.0003cm}
\end{figure*}

\subsubsection{Statistical analysis}

Once the error values for each method over each dataset have been recorded,  a statistical analysis will be carried out. This analysis allows us to compare correctly the performance of the different deep-learning architectures.  Since we are studying multiple models over multiple datasets, the Friedman test is the recommended method  \cite{Garcia:2008}. This non-parametric test allows detecting global differences and provides a ranking of the different methods.  In the case of obtaining a p-value below the significance level (0.05), the null hypothesis (all algorithms are equivalent) can be rejected, and we can proceed with the post-hoc analysis. For this purpose, we use Holm-Bonferroni's procedure, which performs pairwise comparisons between the models. With this $n \times n$ procedure, we can detect significant differences between each pair of models, which allows establishing a statistical ranking.

In this study, we perform the statistical analysis over different performance metrics to evaluate the seven types of deep learning models from all perspectives. We rank the models according to the best results in terms of forecasting accuracy (WAPE) and efficiency (training and inference times). Furthermore, we evaluate the statistical differences of the worst performance and the mean and standard deviation of the results obtained with each type of architecture. Finally, we carry out a ranking of average rankings to see which models are better overall.

An additional statistical analysis is performed using a paired Wilcoxon signed-rank test, in order to study the statistical difference among the studied architecture configurations of each type of model. Note that in this case, we compare models within the same type of deep learning network, so we will discover what configurations are optimal for each of them. The null hypothesis is that models with two different values for a specific configuration are not statistically different at the significance level of $\alpha = 0.05$.  A p-value $\leq$ 0.05 rejects the null hypothesis, indicating a significant difference between the models \cite{Wilcoxon:1992}.

\section{Results and Discussion}
\label{results}

This section reports and discusses the results obtained from the experiments carried out. It provides an analysis of the results in terms of forecasting accuracy and computational time.  The experiments have taken around four months, using five NVIDIA TITAN Xp 12GB GPU in computers with an Intel i7-8700 CPU, and an additional GPU in the Amazon cloud service.

\begin{table*}[t]
\begin{center}
\tbl{Best WAPE results obtained with each type of architecture for all datasets.
\label{tab:top-results}}{
\resizebox{1\textwidth}{!}{%
\begin{tabular}{c||cccccccccccc}
\hline
 \backslashbox{\textbf{Model}}{\textbf{Dataset}} & \textbf{1} & \textbf{2} & \textbf{3} & \textbf{4} & \textbf{5} & \textbf{6} & \textbf{7} & \textbf{8} & \textbf{9} & \textbf{10} & \textbf{11} & \textbf{12} \\ \hline \hline
\textbf{MLP} & 14.886 & 21.245 & 0.367 & 21.114 & 17.191 & 22.930 & 30.193 & 21.365 & 45.515 & 4.373 & 1.891 & 59.710 \\
\textbf{ERNN} & 12.303 & 18.101 & 0.298 & 15.621 & \textbf{14.178} & 18.997 & 27.407 & \textbf{17.958} & 33.354 & 3.388 & 1.374 & 46.739 \\
\textbf{LSTM} & 12.475 & \textbf{15.352} & 0.300 & 15.282 & 14.281 & 18.589 & 24.333 & 19.081 & \textbf{31.960} & 3.359 & \textbf{1.314} & \textbf{46.477} \\
\textbf{GRU} & \textbf{11.596} & 16.772 & 0.314 & \textbf{15.182} & 14.298 & 18.682 & 25.054 & 18.443 & 33.306 & 3.363 & 1.322 & 46.682 \\
\textbf{ESN} & 12.552 & 17.227 & \textbf{0.283} & 17.184 & 14.366 & 19.626 & 25.490 & 18.232 & 38.488 & 3.590 & 1.459 & 47.149 \\
\textbf{CNN} & 12.479 & 17.143 & 0.303 & 15.612 & 14.256 & 18.852 & 29.350 & 18.497 & 34.406 & \textbf{3.337} & 1.333 & 46.914 \\
\textbf{TCN} & 12.866 & 19.091 & 0.303 & 15.587 & 14.575 & \textbf{18.528} & \textbf{23.930} & 18.893 & 32.927 & 3.398 & 1.320 & 46.556 \\ \hline
\end{tabular}%
}}
\end{center}
\end{table*}

\subsection{Forecasting accuracy}

The first part of the analysis is focused on the forecasting accuracy obtained for each model, using the WAPE metric. In Figure \ref{fig:boxplot-wape}, we present a comparison between the distribution of results obtained with all the different model architectures for each dataset.  In some plots, the ERNN distribution has been cut to allow better visualization of the rest of the models. It can be seen at first glance, that some architectures, such as the ERNN or TCN, are more sensitive to the parametrization as they present a wider WAPE distribution compared to others like CNN or MLP.  In general, except for the MLP which is not specifically designed to deal with time series, the rest of the architectures can obtain forecasting accuracy similar to the best model in almost all cases. This can be seen by observing the minimum WAPE values of the models, which are close to each other. However, in this plot, it is more important to analyze how hard it is to achieve such performance. Wider distributions indicate a higher difficulty to find the optimal hyperparameter configuration. In this sense, as will be seen later in the statistical analysis, CNN and LSTM are the most suitable alternatives for a fast design of models with good performance.

Furthermore, Table \ref{tab:top-results} presents a more detailed view of the best results obtained for each architecture on each dataset. As expected, MLP models perform the worst overall. MLP networks are simple models that can serve as a useful comparison baseline with the rest of the architectures. We can notice that LSTM models achieve the best results in four out of twelve datasets and it is in the top three architectures for the remaining datasets, except Tourism for which LSTM is in sixth place only over MLP.  Furthermore, we can also point out that GRU seems to be a very consistent technique as it obtains optimal predictions for most of the datasets, being within the first three architectures for ten out of twelve datasets. TCN and ERNN models obtain the best results in two datasets, similarly to GRU. However,  these two architectures are much more unstable, observing very different results depending on the dataset.  The CNN presents a behavior slightly worse than the TCN in terms of best results. Nevertheless, as it was seen in the boxplot, it outperforms TCN models when comparing the average WAPE for all hyperparameter configurations. Finally, the ESN is one of the worst models, ranking sixth in most datasets, only outperforming MLP.

\begin{figure*}[!t]
	\begin{center}
    \includegraphics[width=0.4\linewidth]{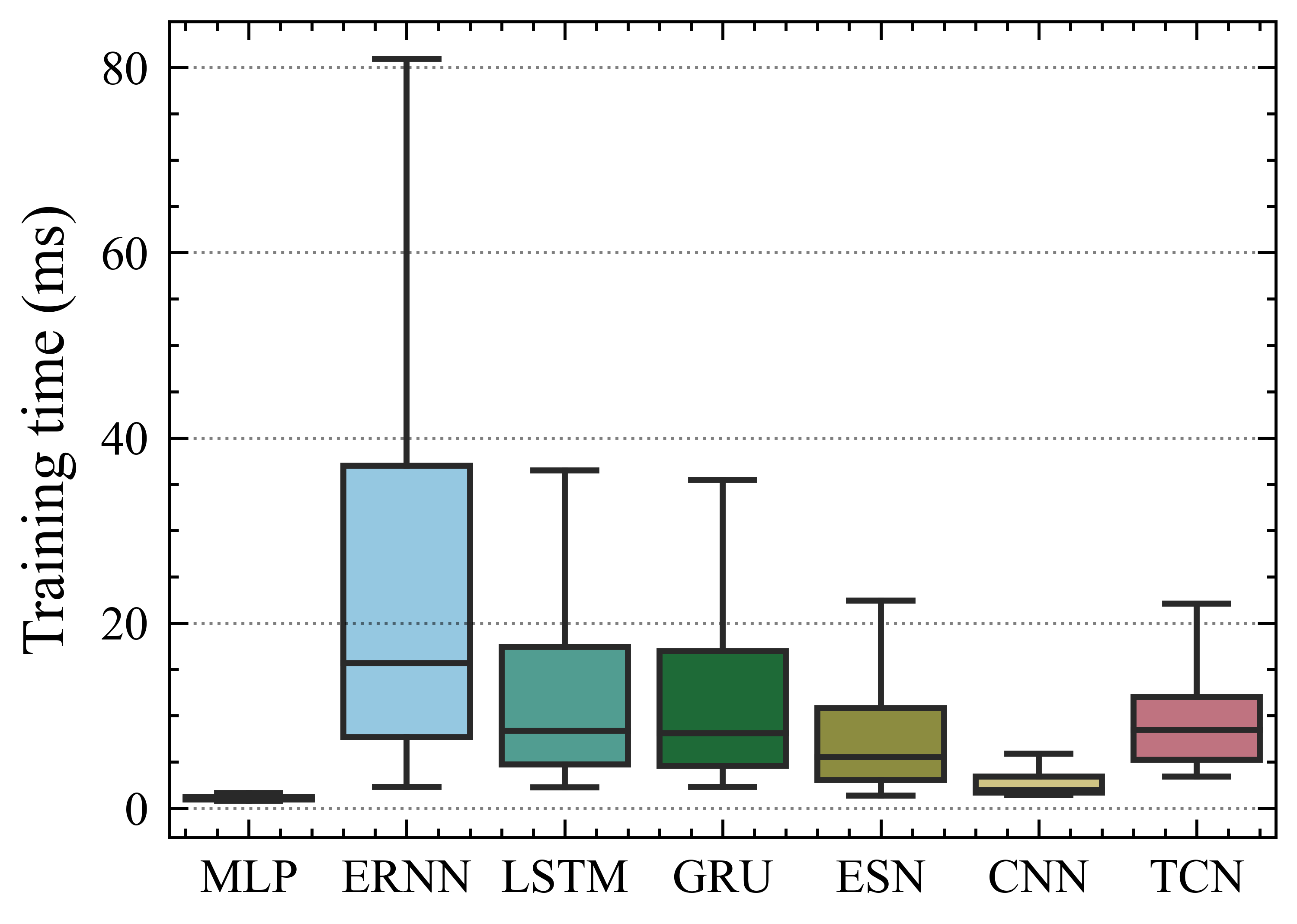}
    \includegraphics[width=0.4\linewidth]{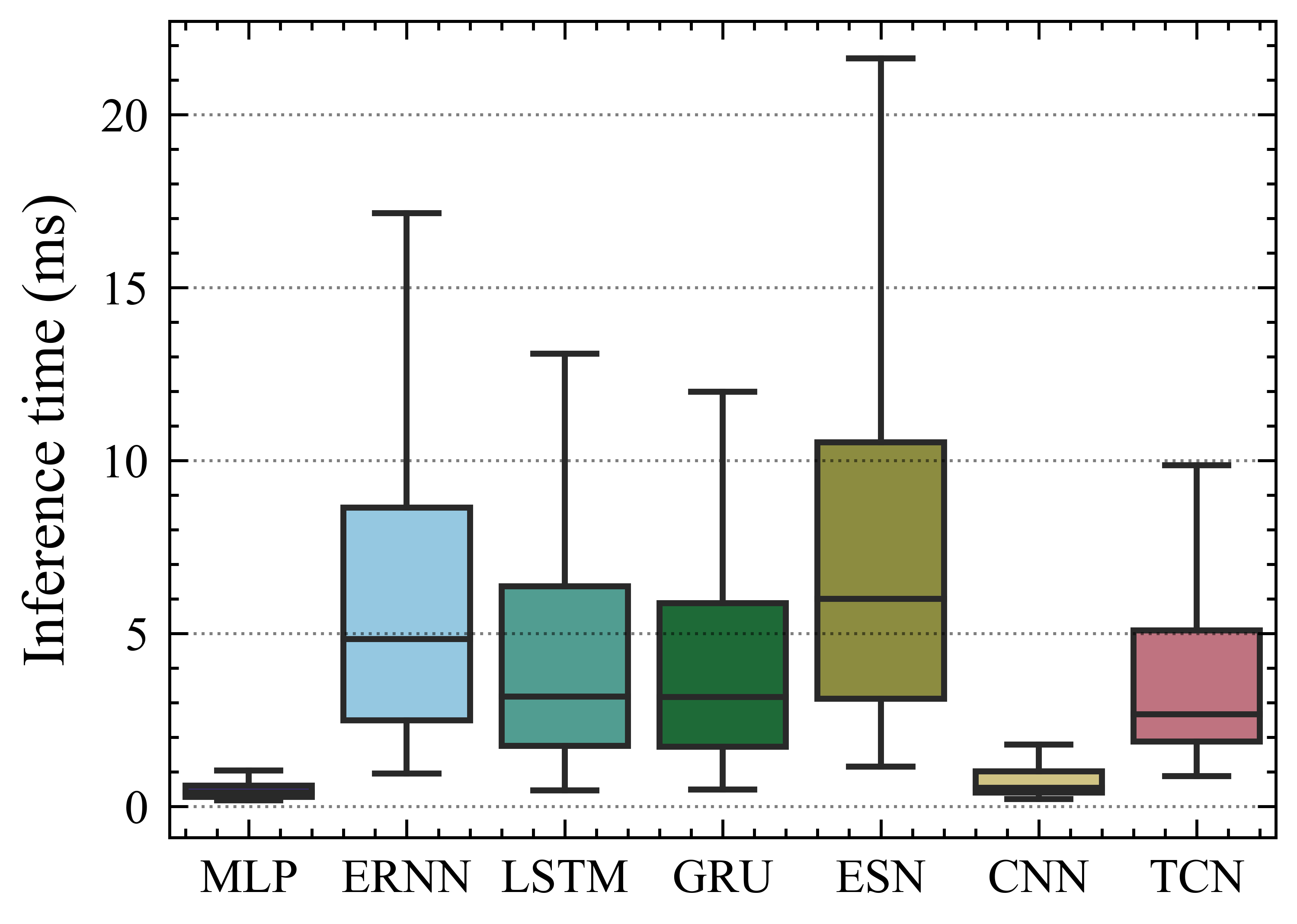}
    \end{center}
    \caption{Distribution of training and inference time results for all architectures.}
    \label{fig:TimeResult}
\end{figure*}

Due to space constraints, the complete report of results is provided in an online appendix that can be found at Ref. \refcite{github-code}. This appendix contains the results for each architecture configuration in several spreadsheets. Furthermore, it has a summary file with the best, mean, standard deviation, and worst results grouped by type of model. It is worth mentioning that CNN outperforms the rest of the models in terms of the mean and standard deviation of WAPE on many datasets. This indicates that it is easier to find a high performing set of parameters for convolutional models than for recurrent ones. However, TCNs have a higher standard deviation of WAPE, given that they are more complex architectures with more parameters to take into account. LSTMs also achieve good performance in terms of average WAPE, which further supports that it is the best alternative among the studied recurrent networks. GRU and ERNN are the models amongst all that suffer a higher standard deviation of results, which demonstrates the difficulty of their hyperparameter tuning. The ESN models also have higher variability of accuracy and do not perform well on average.

\subsection{Computation time}

The second aspect in which the deep learning architectures are evaluated is computational efficiency. Figure  \ref{fig:TimeResult} represents the distribution of training and inference time for each architecture. In general, we can see that MLP is the fastest model, closely followed by CNN. The difference between CNN and the rest of the models is highly significant. Within the recurrent neural networks, LSTM and GRU perform similarly while ERNN is the slowest architecture overall. In general, the analysis for all architectures is analogous when comparing training and inference times. However, the ESN models do not meet this standard as most of the weights of the network are non-trainable, which results in efficient training and slow inference time. Finally, the TCN architecture presents comparable results in terms of average train and inference time with GRU and LSTM models. However, it can be seen that TCN models have less variability in the distribution of computation time. This indicates that a deeper architecture with more number of layers has a smaller effect in convolutional models than in recurrent architectures. Therefore, designing very deep recurrent models may not be suitable under hard time constraints.

\begin{figure}[H]
    \centering
    \includegraphics[width=0.9\linewidth]{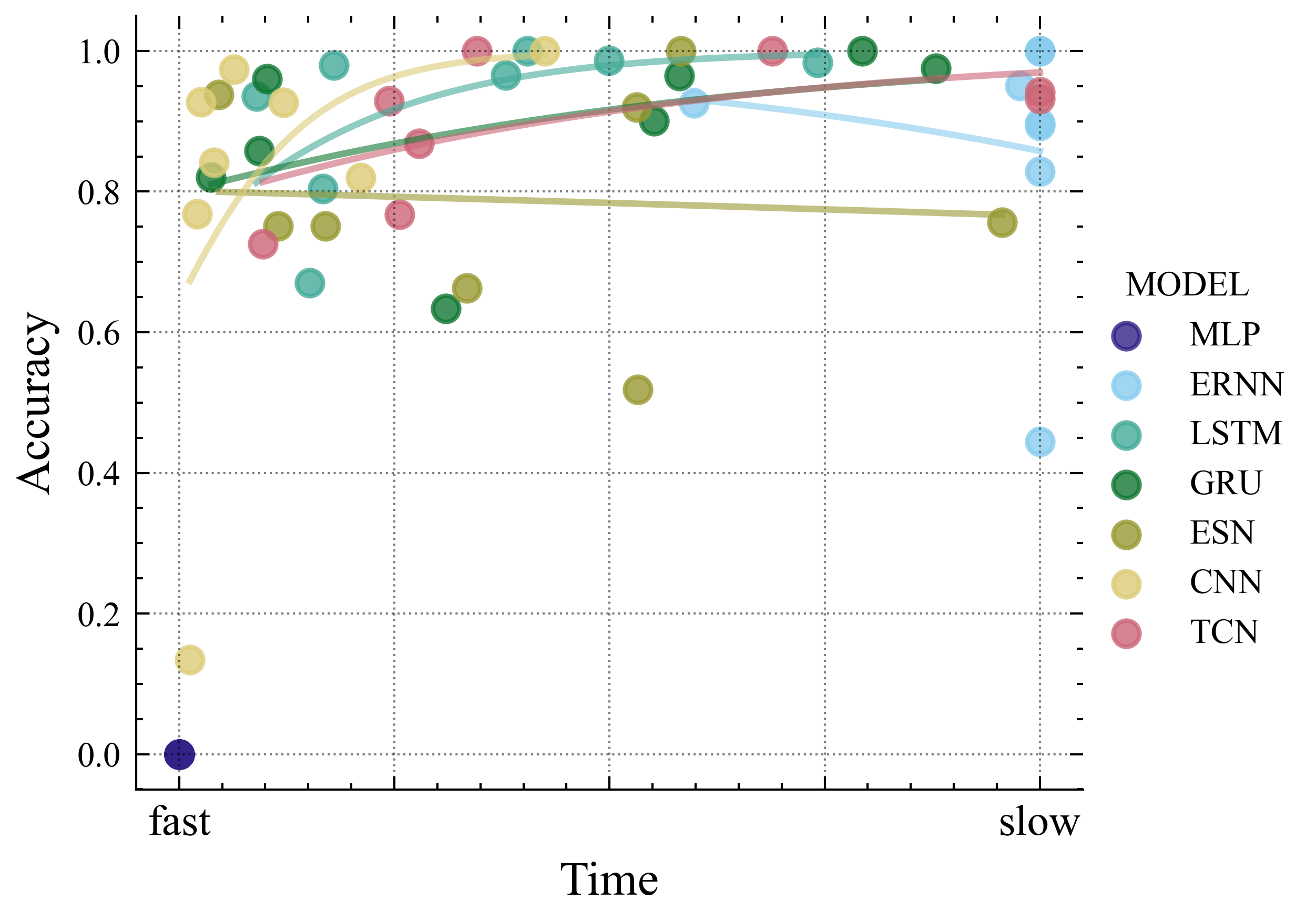}
    \caption{Normalized accuracy versus normalized training time of the best models of each deep learning architecture for each dataset. The lines represent a logistic regression for each model.}
    \label{fig:AccVsTime}
\end{figure}

Figure \ref{fig:AccVsTime} aims to illustrate the speed/accuracy trade-off in this experimental study. The plot presents accuracy against computation time for each deep learning model over each dataset, which confirms the findings discussed previously. The figure shows that MLP networks are the fastest model but provide the worst accuracy results. It can also be seen that the ERNN obtains accurate forecasts but requires high computing resources. Among the recurrent networks, LSTM is the best network with very high predictive performance and adequate computation time. CNN strikes the best time-accuracy balance among all models, with TCN being significantly slower.

\subsection{Statistical analysis}
\label{results:statisical}
\begin{figure*}[!t]
	\begin{center}
    \includegraphics[width=0.83\textwidth]{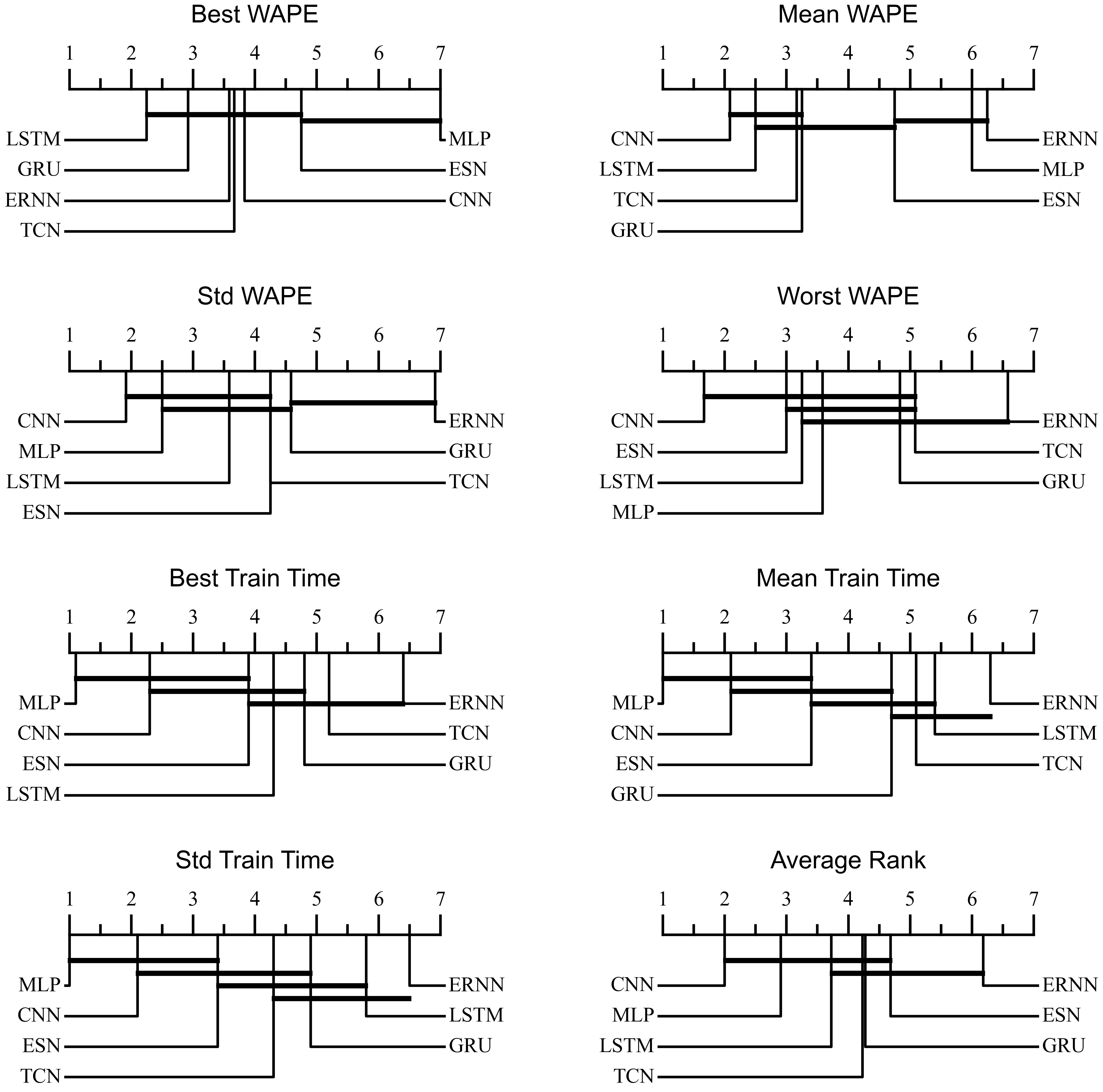}
    \caption{Rankings and critical differences diagram (using the Holm-Bonferroni's post-hoc procedure) of the different deep learning architectures according to several performance metrics}
    \label{fig:cd-diagram}
    \end{center}
\end{figure*}

We perform the statistical analysis, grouping the results obtained with each type of architecture, with respect to eleven different rankings: best WAPE, mean WAPE, WAPE standard deviation, worst WAPE, training and inference time of best model, training and inference mean time, standard deviation of training and inference time, and a global ranking of the average rankings of all these comparisons.
In all cases, the p-value obtained in the Friedman test indicated that the global differences in rankings between architectures  were significant. With these results, we can proceed with the Holm-Bonferroni's post-hoc analysis to perform a pairwise comparison between the models. 
Figure \ref{fig:cd-diagram} presents a compact visualization of the results of the statistical analysis carried out. It displays the ranking of models from left to right for all the considered metrics. Furthermore, it also allows visualizing the significance of the observed paired differences with the plotted horizontal lines. In this critical differences (CD) diagram, models are linked when the null hypothesis of their equivalence is not rejected by the test. The plots in which there are several over-lapped groups indicate that it is hard to detect differences between models with similar performance.

In terms of the best WAPE forecasting accuracy, the recurrent networks lead the ranking. LSTM ranks first, providing the best precision for most of the datasets, closely followed by GRU. The convolutional models, TCN and CNN, obtain rank four and five respectively. However, the CD diagram tells that the statistical differences among the best models obtained for each type of architecture are not significant, except for the MLP. This fact indicates that, when the optimal hyperparameter configuration is found, all these architectures can achieve similar performance for TSF.

When analyzing the mean WAPE ranking, the results are completely different. In this case, CNN ranks first, with LSTM in second place. This indicates that these two models are less sensitive to the parametrization, hence being the best alternatives to reduce the high cost of performing an extensive grid search. This fact is further supported by the WAPE standard deviation diagram, in which CNN and LSTM also occupy the first positions. The MLP models logically have a low standard deviation given their simplicity. The difference in ranking between CNN and LSTM is greater in the standard deviation, confirming that CNN performs well under a wider range of configurations. We also notice that ERNN provides the worst performance in terms of the distribution of results, even below the MLP. This suggests that the parametrization of ERNNs is complex, hence being the less recommended method amongst all the studied.

With regard to the efficiency, the conclusions obtained analyzing training and inference times were similar in both cases. Therefore, to simplify the visualization, we only display the rankings referring to the training time. 
Despite its poorer forecasting accuracy, the MLP ranks first in computational efficiency. MLP networks are the fastest models and also present a very low training-time standard deviation. CNN models appear as the most efficient DL model in all rankings, with ESN in third place. It can also be seen that there are differences between the first time diagram (training time of the model with best performing hyperparameter configuration) and the second (mean training time among all configurations). LSTM has a much lower rank in terms of average computation time and standard deviation, which indicates that designing very deep LSTM models is very costly and not convenient for real-time applications.  Furthermore, it is worth mentioning that CNN seems a better option than TCN for univariate TSF. Although TCNs are better designed for time series, CNN models have shown to provide similar performance in terms of best WAPE results. Furthermore, CNNs have shown to be more computationally efficient and  simpler in terms of finding a good hyperparameter configuration.

With all these analyses in mind, it can be concluded that CNN and LSTM are the most suitable alternatives to approach these TSF problems, as it is displayed in the last diagram (ranking of average rankings). While LSTM provides the best forecasting accuracy, CNNs are more efficient and suffer less variability of results. This accuracy versus efficiency trade-off implies that the most convenient model will be dependent on the problem requirements, time constraints, and the objective of the researcher.

\subsection{Model architecture configuration}

Designing the most appropriate architecture for the studied deep learning models is a very complex task that requires considerable expertise. Therefore, for each architecture, we analyze the results obtained with the different parameters in terms of the number of layers and neurons, the characteristics of convolutional and recurrent units, etc. We present the accuracy results tables with the architecture configurations ordered by WAPE average rankings. Note that, in this analysis, the rankings are independent for each of the seven types of networks. Each model configuration is trained several times with different training hyper-parameters. Given the large number of possibilities, for simplicity, we only report the best WAPE results for each configuration. Among these best results, we display in the tables the four top-ranked and the three worst configurations (among the best WAPE) for each type of model. The complete results tables are provided in the online appendix \cite{github-code}.

\begin{figure*}[!t]
    \centering
     \includegraphics[width=0.32\linewidth]{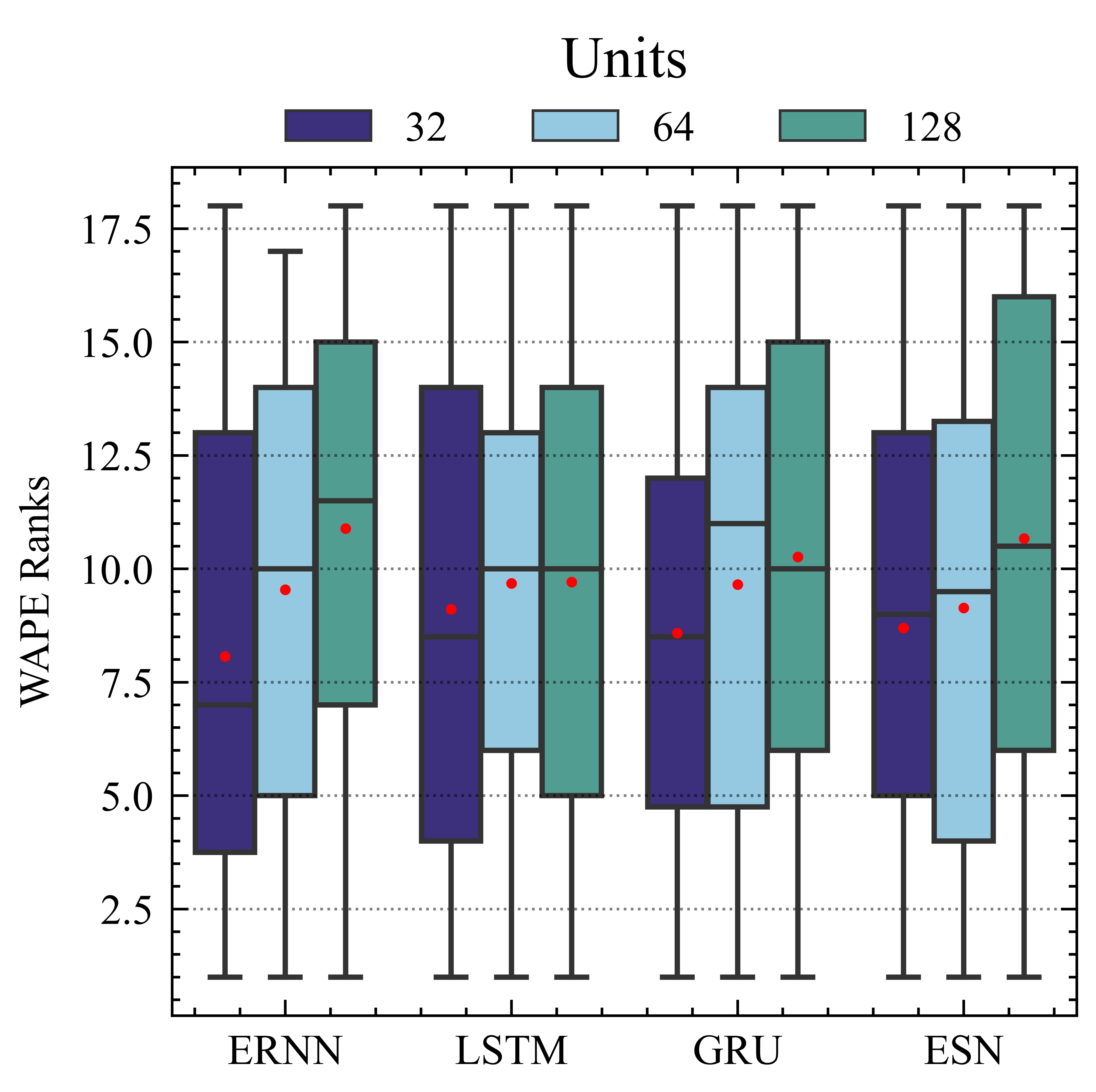}
    \includegraphics[width=0.32\linewidth]{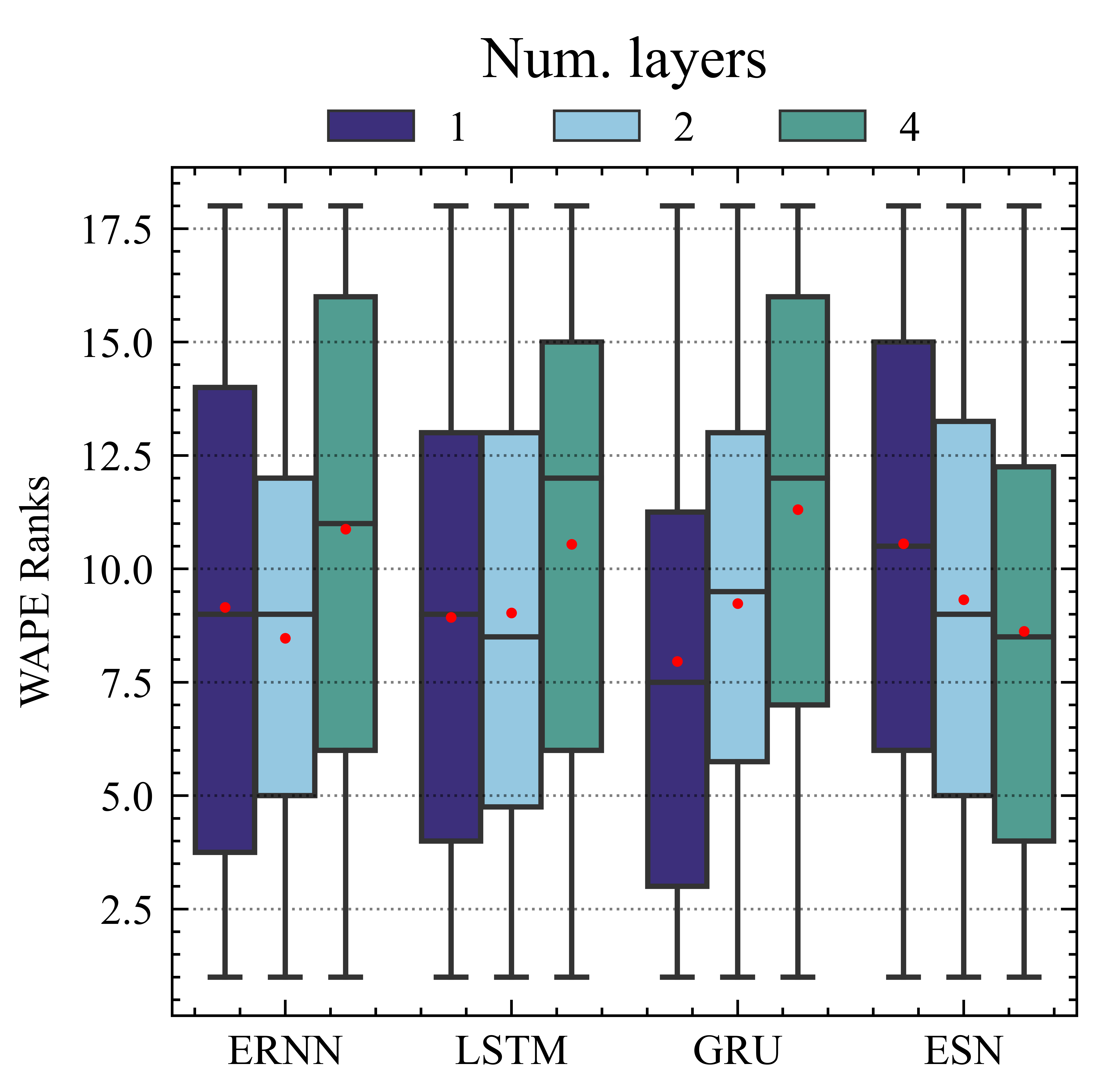}
    \includegraphics[width=0.32\linewidth]{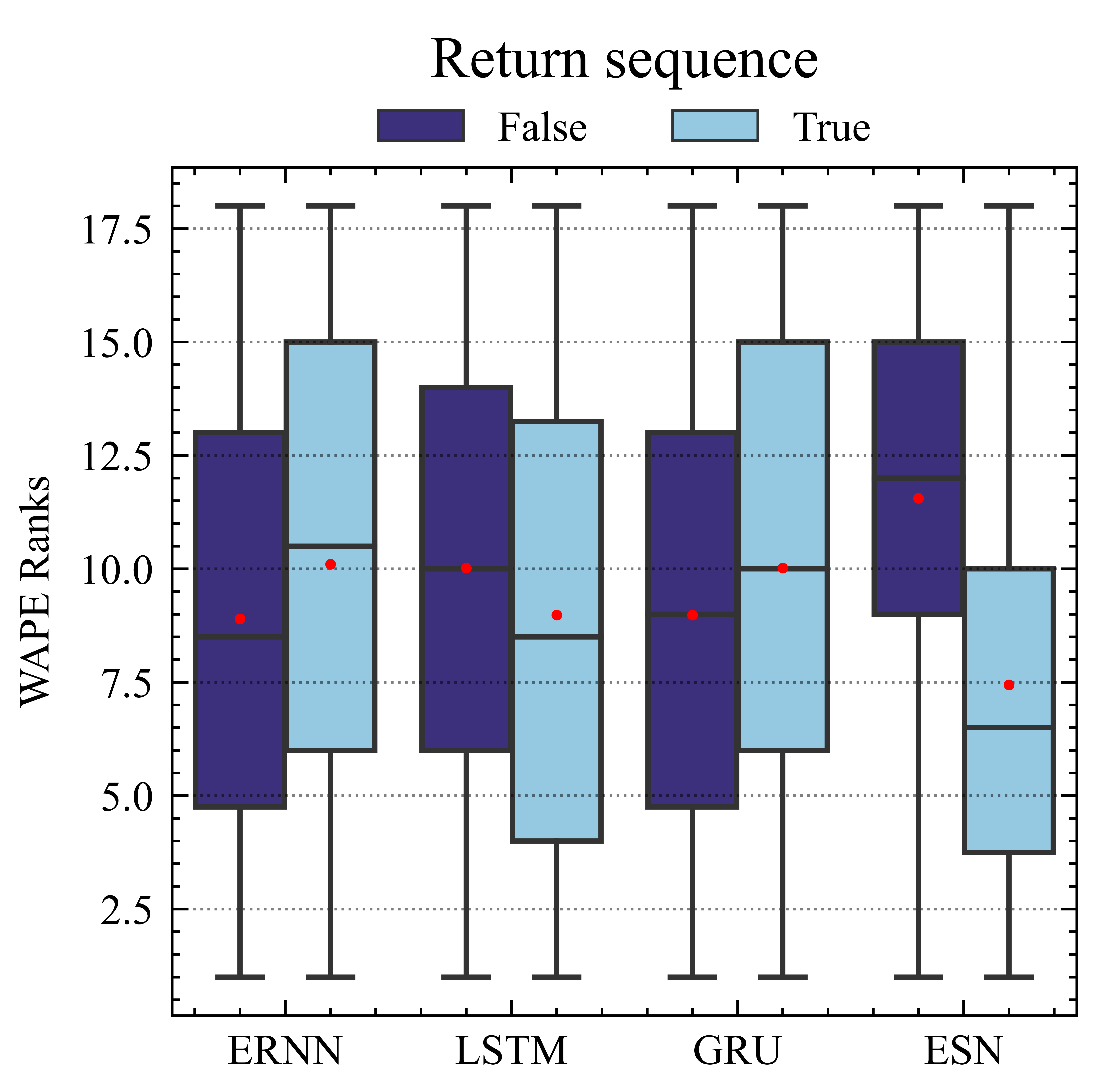}
    \caption{Distribution of WAPE ranks obtained for each recurrent architecture with the different configurations studied. The red dot represents the mean average. }
    \label{fig:rnn-params}
\end{figure*}

\subsubsection{Multi-Layer Perceptron (MLP)}

Table \ref{tab:results-rank-mlp} presents the best results for each configuration of the MLP networks. The top-ranked MLP model is composed of 5 hidden layers, with a total of 80 neurons. The results in the table confirm the findings that can be read in the literature, given that the design of MLP is a well-studied field. They show that the increase of hidden neurons does not imply better performance. In fact, the most complex model, an MLP with 320 neurons distributed in 5 layers, is positioned penultimate, obtaining worse predictions than a simple network of one layer of 8 neurons.

\begin{table}[H]
\tbl{WAPE results of the best MLP architecture configurations.
\label{tab:results-rank-mlp}}{
\resizebox{0.65\linewidth}{!}{%
\begin{tabular}{c||c||c}
\hline
\textbf{\#} &   & \textbf{Mean ranks} \\ \hline\hline
 & \textbf{Hidden layers}  &
\\
1 & {[}8, 16, 32, 16, 8{]} & \textbf{5.417} \\
2 & {[}32, 64{]} & 5.917 \\
3 & {[}128, 64, 32{]} & 6.167 \\
 ··· &  ···  & ···· \\
10 & {[}16, 8{]} & 7.000 \\
11 & {[}32, 64, 128, 64, 32{]} & 7.083 \\
12 & {[}32{]} & 7.583 \\ \hline
\end{tabular}%
}}
\end{table}

\subsubsection{Recurrent Neural Networks}
Among the recurrent architectures, the best performance was obtained with LSTM models. Table \ref{tab:results-lstm} presents the WAPE metrics obtained for the different LSTM model configurations. The best results were obtained with two stacked layers of 32 units which returns the complete sequence before the output dense layer.

Since all recurrent networks have the same parameters, we present a global comparison in Figure \ref{fig:rnn-params}. This plot shows the results of the different RNN architectures depending on the three main parameters: number of layers, units, and whether sequences are returned. On average, we can observe that all recurrent models tend to obtain better predictions with a small number of units. When analyzing the number of stacked layers parameter, we can observe that LSTM, ERNN, and GRU have similar behavior, obtaining worse average results when it is increased. However, ESN models present the opposite pattern, with the best results obtained with 4 layers. With regard to returning the complete sequence or just the last output, it can clearly be seen that ESN models achieve better performance when returning the whole sequence.

\begin{table}[H]
\tbl{WAPE results of the best LSTM architecture configurations.
\label{tab:results-lstm}}{
\resizebox{0.9\linewidth}{!}{%
\begin{tabular}{c||ccc||c}
\hline 
\textbf{\#} &   & \textbf{Model} &  & Mean ranks \\ \hline\hline
 & \textbf{Num. Layers} & \textbf{Units} & \textbf{Return sequence} &  \\
1 & 2 & 32 & True & \textbf{6.667} \\
2 & 2 & 64 & True & 7.500 \\
3 & 1 & 32 & False & 8.083 \\
···&  & · · · & &  · · · \\
16 & 4 & 128 & False & 11.250\\
17 & 2 & 128 & False & 11.500\\
18 & 4 & 64 & False & 11.917\\ \hline
\end{tabular}
}}
\end{table}

\subsubsection{Convolutional Neural Networks}

Table \ref{tab:results-rank-cnn} reports the ranking of the best CNN model configurations. It can be noticed that the prediction accuracy is proportional to the number of convolutional layers. The best results have been obtained with models with four layers while the single-layer models are at the bottom of the ranking. However, regarding the number of filters, there is no significant difference at first sight. It is also worth mentioning that the best predictions have been obtained from models without max-pooling, suggesting that this popular image-processing operation is not suitable for time series forecasting.

\begin{table}[H]
\tbl{WAPE results of the best CNN architecture configurations.
\label{tab:results-rank-cnn}}{
\resizebox{0.85\linewidth}{!}{%
\begin{tabular}{c||ccc||c}
\hline
 \textbf{\#} & & \textbf{Model} &  & \textbf{Mean ranks} \\ \hline\hline
 & \textbf{N. Layers} & \textbf{N. Filters} & \textbf{Pool factor} & \\
1 & 4 & 16 & 0 & \textbf{6.000} \\
2 & 4 & 64 & 0 & 6.750 \\
3 & 4 & 64 & 2 & 6.917 \\
···&  & · · · & &  · · · \\
16 & 1 & 32 & 2 & 11.917 \\
17 & 1 & 64 & 2 & 12.000 \\
18 & 1 & 16 & 0 & 13.000 \\ \hline
\end{tabular}%
}}
\end{table}

\begin{table}[H]
\tbl{WAPE results of the best TCN architecture configurations.
\label{tab:results-rank-tcn}}{
\resizebox{\linewidth}{!}{%
\begin{tabular}{c||ccccc||c}
\hline 
  \textbf{\#} & & & \textbf{Model} & &  &  \textbf{Mean ranks} \\ \hline\hline
 & \textbf{N. Layers} & \textbf{N. Filters} & \textbf{Dilations} & \textbf{Kernel} & \textbf{Return seq.} &   \\
1 & 1 & 64 & {[}1, 2, 4, 8{]} & 6 & False & \textbf{11.083} \\
2 & 1 & 32 & {[}1, 2, 4, 8{]} & 6 & False & 11.500 \\
3 & 4 & 32 & {[}1, 2, 4, 8, 16{]} & 6 & False & 11.500 \\
···&  & & · · · & & &  · · · \\
30 & 4 & 64 & {[}1, 2, 4, 8{]} & 3 & False & 21.250 \\
31 & 4 & 64 & {[}1, 2, 4, 8, 16{]} & 3 & True & 21.500 \\
32 & 4 & 64 & {[}1, 2, 4, 8, 16{]} & 3 & False & 21.667 \\ \hline
\end{tabular}}}
\end{table}


\begin{table*}[!t]
\tbl{Example of how LSTM results are gathered to compare the number of layers parameter with the Wilcoxon signed-ranked test.
\label{tab:wilcoxon-example}}{
\resizebox{0.85\textwidth}{!}{%
\begin{tabular}{c||ccc||ccccccccc}
\hline
\textbf{\#} &  \multicolumn{3}{c||}{\backslashbox{Parameters}{{Fixed parameter}}} & \multicolumn{9}{c}{\textbf{Layers}} \\ \hline\hline
& \textbf{Units} & \textbf{Return Seq.} & \textbf{Dataset} & 1 & vs & \multicolumn{1}{c|}{2} & 1 & vs & \multicolumn{1}{c|}{4} & 2 & vs & 4 \\
\textbf{1} & 32 & False & 1 & 14.554 & - & \multicolumn{1}{c|}{15.224} & 14.554 & - & \multicolumn{1}{c|}{14.747} & 15.224 & - & 14.747 \\
\textbf{2} & 32 & False & 2 & 16.212 & - & \multicolumn{1}{c|}{15.352} & 16.212 & - & \multicolumn{1}{c|}{22.489} & 15.352 & - & 22.489 \\
\textbf{3} & 32 & False & 3 & 0.312 & - & \multicolumn{1}{c|}{0.349} & 0.312 & - & \multicolumn{1}{c|}{0.337} & 0.349 & - & 0.337 \\
\textbf{···} &  & ··· &  &  & ··· & \multicolumn{1}{c|}{} &  & ··· & \multicolumn{1}{c|}{} &  & ··· &  \\
\textbf{12} & 32 & False & 12 & 46.477 & - & \multicolumn{1}{c|}{46.714} & 46.477 & - & \multicolumn{1}{c|}{47.020} & 46.714 & - & 47.020 \\
\textbf{13} & 32 & True & 1 & 13.372 & - & \multicolumn{1}{c|}{12.749} & 13.372 & - & \multicolumn{1}{c|}{13.672} & 12.749 & - & 13.672 \\
\textbf{14} & 32 & True & 2 & 18.161 & - & \multicolumn{1}{c|}{20.193} & 18.161 & - & \multicolumn{1}{c|}{28.026} & 20.193 & - & 28.026 \\
··· &  & ··· &  &  &  & \multicolumn{1}{c|}{} &  &  & \multicolumn{1}{c|}{} &  &  &  \\
\textbf{24} & 32 & True & 12 & 46.608 & - & \multicolumn{1}{c|}{46.758} & 46.608 & - & \multicolumn{1}{c|}{47.067} & 46.758 & - & 47.067 \\
\textbf{25} & 64 & False & 1 & 14.259 & - & \multicolumn{1}{c|}{14.150} & 14.259 & - & \multicolumn{1}{c|}{14.419} & 14.150 & - & 14.419 \\
\textbf{26} & 64 & False & 2 & 21.601 & - & \multicolumn{1}{c|}{18.131} & 21.601 & - & \multicolumn{1}{c|}{22.234} & 18.131 & - & 22.234 \\
··· &  & ··· &  &  &  & \multicolumn{1}{c|}{} &  &  & \multicolumn{1}{c|}{} &  &  &  \\
\textbf{48} & 64 & True & 12 & 46.863 & - & \multicolumn{1}{c|}{47.108} & 46.863 & - & \multicolumn{1}{c|}{47.167} & 47.108 & - & 47.167 \\
\textbf{49} & 128 & False & 1 & 14.825 & - & \multicolumn{1}{c|}{14.549} & 14.825 & - & \multicolumn{1}{c|}{15.479} & 14.549 & - & 15.479 \\
··· &  & ··· &  &  &  & \multicolumn{1}{c|}{} &  &  & \multicolumn{1}{c|}{} &  &  &  \\
\textbf{71} & 128 & True & 11 & 1.314 & - & \multicolumn{1}{c|}{1.340} & 1.314 & - & \multicolumn{1}{c|}{1.350} & 1.340 &  & 1.350 \\
\textbf{72} & 128 & True & 12 & 46.822 & - & \multicolumn{1}{c|}{46.858} & 46.822 & - & \multicolumn{1}{c|}{46.994} & 46.858 &  & 46.994 \\ \hline
\end{tabular}%
}}
\end{table*}

Table \ref{tab:results-rank-tcn} shows results obtained with the different configurations of TCN models. The best network overall is designed with one single TCN layer with 4 dilated convolutional layers, a kernel of length 6, and 64 filters that do not return the complete sequence before the output layer. These results indicate a clear pattern for designing TCN models. Firstly, it should be noted that single-layer models with few dilated layers outperform more-complex models. Concerning the kernel size, the models with a
larger kernel size provide the best predictions. Furthermore, according to the results, it is revealed that returning the  complete sequence before the last dense layer may increases the complexity of the model too much, resulting in a performance downturn.

\subsection{Statistical comparison of architecture configurations}

Using the results presented in the previous subsection, we present a statistical analysis comparing the studied architecture configurations for each type of model. This study aims to confirm the findings that were discussed previously regarding the best parameter choices. The method used for this comparison is the paired Wilcoxon signed-rank test. Table \ref{tab:wilcoxon-example} illustrates an example of how results are gathered for the comparison. Given a certain parameter, we compare the results between each pair of possible values, keeping fixed the rest of the parameters. It is worth mentioning that, given the large sample distribution of possible architecture configurations, the results presented here are both reliable and significant. For instance, in the RNN case, a total of 108 samples are compared for parameters with 2 possible values, while 72 samples are compared for parameters with 3 values.

Table \ref{tab:optimal-architectures} presents the findings obtained from the Wilcoxon test, providing the best configurations found for each type of architecture. The numbers with ** indicate that it is the best value with a significant statistical difference compared to the rest $(p < 0.05)$. We indicate with * that there is a certain tendency suggesting that it is better to choose that parameter $(p < 0.2)$, and with $=$ when there are no significant differences between choosing any of the possible parameters.

\begin{table}[H]
\tbl{Best architecture configuration for each type of model. Values with ** are significantly better $(\textit{p-value} \leq 0.05)$, * suggests it is a better choice although not significant $(\textit{p-value} \leq 0.2)$, and $=$ means that no differences were found among the possible values.
\label{tab:optimal-architectures}}{
\resizebox{0.95\linewidth}{!}{%
\begin{tabular}{cc||cc}
\hline
\multicolumn{2}{c||}{\textbf{ERNN}} & \multicolumn{2}{c}{\textbf{LSTM}}  \\ \hline
Layers & 1**, 2* & Layers & 1*, 2**  \\
Units & 32**, 64* & Units & =  \\
\multirow{1}{*}{Return Sequence} & \multirow{1}{*}{False*} & \multirow{1}{*}{Return Sequence} & \multirow{1}{*}{=}  \\ \hline \hline
 
\multicolumn{2}{c||}{\textbf{GRU}} & \multicolumn{2}{c}{\textbf{ESN}} \\  \hline
 Layers & \multicolumn{1}{c||}{1**, 2**} & Layers & 2*, 4** \\
 Units & \multicolumn{1}{c||}{32*} &  Units & 32**, 64**\\
 Return sequence & = & Return Sequence & True**\\\hline \hline
 
 \multicolumn{2}{c||}{\textbf{CNN}} & \multicolumn{2}{c}{\textbf{TCN}} \\ \hline
 Layers & 4** & Layers & 1** \\
 \multirow{2}{*}{Filters} & \multirow{2}{*}{=} & Filters & 32* \\
   &  & Dilations & = \\
 \multirow{2}{*}{Pooling factor} & \multirow{2}{*}{0**} & Kernel & 6** \\
   &  & Return Sequence & True* \\ \hline
\end{tabular}%
}}
\end{table}

\begin{table*}[!t]
\tbl{Best training hyperparameter values for each architecture. Values with ** are significantly better $(\textit{p-value}\leq 0.05)$, * suggests it is a better choice although not statistically significant $(\textit{p-value} \leq 0.2)$, and $=$ means that no differences were found among the possible values.
\label{tab:optimal-training-params}}{
\resizebox{0.9\linewidth}{!}{%

\begin{tabular}{c||ccccccc}
\hline
 \backslashbox{\textbf{Parameter}}{\textbf{Architecture}} & \textbf{MLP} & \textbf{ERNN} & \textbf{LSTM} & \textbf{GRU} & \textbf{ESN} & \textbf{CNN} & \textbf{TCN} \\ \hline\hline
\textbf{Batch size} & = & 64** & 32** & 64** & = & 64** & 64* \\
\textbf{Past History factor} & 1.25** & 1.25** & 1.25** & 1.25** & 1.25** & 1.25** & 1.25** \\
\textbf{Learning Rate} & 0.001** & 0.001** & 0.001** & 0.001** & 0.001** & 0.001** & 0.001** \\
\textbf{Normalization Method} & zscore** & zscore** & minmax** & minmax* & zscore** & zscore** & zscore** \\ \hline
\end{tabular}}
}
\end{table*}

For the recurrent networks, it can be seen that a lower number of stacked layers (one or two) provides better performance. The only exception is the ESN, for which the optimal value is four layers. With regard to the number of units, the values 32 or 64 are better than 128 for ERNN, GRU, and ESN, while no differences are found for the LSTM. Furthermore, the statistical test confirms that ESN achieves significantly better performance when returning the whole sequence. It also suggests that ERNN works better without returning the sequences, while the performance of LSTM and GRU is not affected by this parameter. In the case of LSTM models, it can be seen that the parameter choice has a minimal impact on the accuracy results. This finding further supports the fact that LSTMs are the most robust type of recurrent network. As was seen in Section \ref{results:statisical}, LSTM models were the best in terms of mean WAPE, showing less variability of results. Since no important difference in performance was found among the parameters, finding a high performing architecture design is easier.


For CNN models, the Wilcoxon test confirms that stacking four layers is significantly better than using just one and two. The number of filters does not affect performance, while it is clear that max-pooling is not recommended for these TSF problems. In the case of TCNs, the test indicates that single-layer models outperform more-complex models. Additionally, it is also confirmed that the models with a larger kernel size provide significantly better predictions. Concerning the number of filters and dilated layers, no statistical difference was found. Furthermore, the results suggest that returning the complete sequence before the last dense layer may increase the complexity of the model excessively, resulting in performance degradation.

\subsection{Analysis of the training parameters}

The grid search performed in the experimental study also involved several training hyperparameters. 
We use the paired Wilcoxon signed-rank test to be able to compare the results. Table \ref{tab:optimal-training-params} summarizes the best training parameter values for each deep learning architecture. First of all, a clear pattern that is common to all architectures can be noted. The best results are achieved with a low learning rate and a small past history factor. Regarding the normalization method, we can observe that the min-max normalization method performs better for GRU and LSTM while z-score significantly beats the min-max method in the other architectures. Furthermore, results also show that most architectures (ERNN, GRU, CNN, and TCN) perform better with larger batch sizes. Only for LSTM, the preferred value is 32, while no difference was found for MLP and ESN.

\section{Conclusions}
\label{conclusions}

In this paper, we carried out an experimental review on the use of deep learning models for time series forecasting. We reviewed the most successful applications of deep neural networks in recent years, observing that recurrent networks have been the most extended approach in the literature. However, convolutional networks are increasingly gaining popularity due to their efficiency, especially with the development of temporal convolutional networks.

Furthermore, we conducted an extensive experimental study using seven popular deep learning architectures: multilayer perceptron (MLP), Elman recurrent neural network ERNN, long-short term memory (LSTM), gated recurrent unit (GRU), echo state network (ESN), convolutional neural network (CNN) and temporal convolutional network (TCN). We evaluated the performance of these models, in terms of accuracy and efficiency, over 12 different forecasting problems with more than 50000 time series in total. We carried out an exhaustive search of architecture configuration and training hyperparameters, building more than 38000 different models. Moreover, we performed a thorough statistical analysis over several metrics to assess the differences in the performance of the models.

The conclusions obtained from this experimental study are summarized below:
\begin{itemize}
    \item Except MLP, all the studied models obtain accurate predictions when parametrized correctly. However, the distribution of results of the models presented significant differences. This illustrates the importance of finding an optimal architecture configuration.
    \item Regardless of the depth of their hidden blocks, MLP networks are unable to model the temporal order of the time series data, providing poor predictive performance.
    \item LSTM obtained the best WAPE results followed by GRU. However, CNN outperforms them in the mean and standard deviation of WAPE. This indicated that convolutional architectures are easier to parameterize than the recurrent models.
    \item CNNs strike the best speed/accuracy trade-off, which makes them more suitable for real-time applications than recurrent approaches.
    \item LSTM networks obtain better results with a lower number of stacked layers, contrary to the GRU architecture. Within recurrent networks, the number of units was not important and returning the complete sequences only proved useful in the ESN models.
    \item CNNs require more stacked layers to improve accuracy, but without using max-pooling operations. For TCNs, a single-block with a larger kernel size is recommended.
    \item With regard to the training hyperparameters, we discovered that lower values of past history and learning rates are a better choice for these deep learning models. CNNs perform better with z-score normalization, while min-max is suggested for LSTM.

\end{itemize}


In future works, we aim to study the application of these deep learning techniques for time series forecasting in streaming. In this real-time scenario, a more in-depth analysis of the speed versus accuracy trade-off will be necessary. Another relevant work would be to study the best deep learning models depending on the nature of the time-series data and other characteristics such as the length or the forecasting horizon. It would also be important to focus future experiments on tuning the training hyperparameters, once that the most suitable model architectures have been found. Other possible extensions of this study are the analysis of regularization techniques in deep networks and performing the experimental study over multivariate time series. Furthermore, future research should address current trends in the literature such as ensemble models to enhance accuracy or transfer learning to reduce the burden of high training times. Future efforts should also be focused on building a larger high-quality forecasting database that could serve as a general benchmark for validating novel proposals. 


\section*{Funding}{This research has been funded by FEDER/Ministerio de Ciencia, Innovación y Universidades – Agencia Estatal de Investigación/Proyecto TIN2017-88209-C2 and by the Andalusian Regional Government under the projects: BIDASGRI:  Big~Data technologies for Smart Grids (US-1263341), Adaptive hybrid models to predict solar and wind renewable energy production (P18-RT-2778).}

\section*{Acknowledgments}{We are grateful to NVIDIA
for their GPU Grant Program that has provided us high-quality GPU devices for carrying out the study.}

\bibliographystyle{ws-ijns}
\bibliography{bibliography}

\begin{thebibliography}{10}

\bibitem{Zhao:2016}
Y.~Zhao, L.~Ye, Z.~Li, X.~Song, Y.~Lang and J.~Su, A novel bidirectional
  mechanism based on time series model for wind power forecasting, {\em Applied
  Energy} {\bf 177}  (2016)  793 -- 803.

\bibitem{Deb:2017}
C.~Deb, F.~Zhang, J.~Yang, S.~E. Lee and K.~W. Shah, A review on time series
  forecasting techniques for building energy consumption, {\em Renew. and Sust.
  Energ. Rev.} {\bf 74}  (2017)  902 -- 924.

\bibitem{Chen:2015}
M.~Chen and B.~Chen, A hybrid fuzzy time series model based on granular
  computing for stock price forecasting, {\em Inf. Sciences} {\bf 294}  (2015)
  227 -- 241.

\bibitem{rafiei:2015}
M.~H. Rafiei and H.~Adeli, A novel machine learning model for estimation of
  sale prices of real estate units, {\em Journal of Construction Engineering
  and Management} {\bf 142} (08 2015) p. 04015066.

\bibitem{Tuarob:2017}
S.~Tuarob, C.~S. Tucker, S.~Kumara, C.~L. Giles, A.~L. Pincus, D.~E. Conroy and
  N.~Ram, How are you feeling?: A personalized methodology for predicting
  mental states from temporally observable physical and behavioral information,
  {\em J. Biomed. Inform.} {\bf 68}  (2017)  1 -- 19.

\bibitem{Ashok:2018}
M.~Paolanti, D.~Liciotti, R.~Pietrini, A.~Mancini and E.~Frontoni, Online
  detection of stealthy false data injection attacks in power system state
  estimation, {\em IEEE Trans. on Smart Grid} {\bf 9}(3)  (2018)  1636--1646.

\bibitem{Zhang:2019}
Y.~Zhang, T.~Cheng and Y.~Ren, A graph deep learning method for short-term
  traffic forecasting on large road networks, {\em Computer-Aided Civil and
  Infrastructure Engineering} {\bf 34}(10)  (2019)  877--896.

\bibitem{Schmidhuber:2015}
J.~Schmidhuber, Deep learning in neural networks: An overview, {\em Neural
  Networks} {\bf 61}  (2015)  85 -- 117.

\bibitem{Sebastian:2019}
O.~Reyes and S.~Ventura, Performing multi-target regression via a parameter
  sharing-based deep network, {\em International Journal of Neural Systems}
  {\bf 29}(09)  (2019) p. 1950014.

\bibitem{Lara:2020}
P.~Lara-Ben{\'{i}}tez, M.~Carranza-Garc{\'{i}}a,
  J.~Garc{\'{i}}a-Guti{\'{e}}rrez and J.~Riquelme, Asynchronous dual-pipeline
  deep learning framework for online data stream classification, {\em
  Integrated Computer-Aided Engineering} {\bf 27}(2)  (2020)  101--119.

\bibitem{Gamboa:2017}
J.~C.~B. Gamboa, Deep learning for time-series analysis, {\em CoRR} {\bf
  abs/1701.01887}  (2017).

\bibitem{Hewamalage:2019}
H.~Hewamalage, C.~Bergmeir and K.~Bandara, Recurrent neural networks for time
  series forecasting: Current status and future directions, {\em Int. Journal
  of Forecasting}   (2020).

\bibitem{Sezer:2020}
O.~B. Sezer, M.~U. Gudelek and A.~M. Ozbayoglu, Financial time series
  forecasting with deep learning : A systematic literature review: 2005–2019,
  {\em Appl. Soft Comp.} {\bf 90}  (2020) p. 106181.

\bibitem{Hyndman:2008}
R.~Hyndman, A.~Koehler, K.~Ord and R.~Snyder, {\em Forecasting with Exponential
  Smoothing} (Springer Berlin Heidelberg, 2008).

\bibitem{Box:1994}
G.~E.~P. Box and G.~M. Jenkins, {\em Time Series Analysis: Forecasting and
  Control} (Prentice Hall, 1994).

\bibitem{Zhang:2003}
G.~Zhang, Time series forecasting using a hybrid {ARIMA} and neural network
  model, {\em Neurocomputing} {\bf 50}  (2003)  159 -- 175.

\bibitem{Martinez-Alvarez:2015}
F.~Mart{\'{i}}nez-{\'{A}}lvarez, A.~Troncoso, G.~Asencio and J.~Riquelme, {A
  Survey on Data Mining Techniques Applied to Electricity-Related Time Series
  Forecasting}, {\em Energies} {\bf 8}(11)  (2015)  13162--13193.

\bibitem{Raza:2015}
M.~Q. Raza and A.~Khosravi, A review on artificial intelligence based load
  demand forecasting techniques for smart grid and buildings, {\em Renew. and
  Sust. Energ. Rev.} {\bf 50}  (2015)  1352 -- 1372.

\bibitem{Palmer:2006}
A.~Palmer, J.~Montaño and A.~Sesé, Designing an artificial neural network for
  forecasting tourism time series, {\em Tourism Management} {\bf 27}(5)  (2006)
   781 -- 790.

\bibitem{Zhang:2007}
G.~P. {Zhang} and D.~M. {Kline}, Quarterly time-series forecasting with neural
  networks, {\em IEEE Transactions on Neural Networks} {\bf 18}(6)  (2007)
  1800--1814.

\bibitem{Hamzacebi:2009}
C.~Hamzaçebi, D.~Akay and F.~Kutay, Comparison of direct and iterative
  artificial neural network forecast approaches in multi-periodic time series
  forecasting, {\em Expert Syst. Appl.} {\bf 36}(2)  (2009)  3839 -- 3844.

\bibitem{Yan:2012}
W.~{Yan}, Toward automatic time-series forecasting using neural networks, {\em
  IEEE Trans. Neural Netw. Learn. Syst.} {\bf 23}(7)  (2012)  1028--1039.

\bibitem{Claveria:2014}
O.~Claveria and S.~Torra, Forecasting tourism demand to {Catalonia}: Neural
  networks vs. time series models, {\em Economic Modelling} {\bf 36}  (2014)
  220 -- 228.

\bibitem{Jourentzes:2014}
N.~Kourentzes, D.~K. Barrow and S.~F. Crone, Neural network ensemble operators
  for time series forecasting, {\em Expert Syst. Appl.} {\bf 41}(9)  (2014)
  4235 -- 4244.

\bibitem{Rahman:2016}
M.~M. {Rahman}, M.~M. {Islam}, K.~{Murase} and X.~{Yao}, Layered ensemble
  architecture for time series forecasting, {\em IEEE Transactions on
  Cybernetics} {\bf 46}(1)  (2016)  270--283.

\bibitem{Torres:2018}
J.~Torres, A.~Galicia~de Castro, A.~Troncoso and F.~Martínez-Álvarez, A
  scalable approach based on deep learning for big data time series
  forecasting, {\em Integrated Computer-Aided Engineering} {\bf 25} (08 2018)
  1--14.

\bibitem{McCulloch:1943}
W.~S. McCulloch and W.~Pitts, A logical calculus of the ideas immanent in
  nervous activity, {\em The bulletin of mathematical biophysics} {\bf 5}(4)
  (1943)  115--133.

\bibitem{Hippert:2001}
H.~S. {Hippert}, C.~E. {Pedreira} and R.~C. {Souza}, Neural networks for
  short-term load forecasting: a review and evaluation, {\em IEEE Trans. on
  Power Systems} {\bf 16}(1)  (2001)  44--55.

\bibitem{Sfetsos:2000}
A.~Sfetsos and A.~Coonick, Univariate and multivariate forecasting of hourly
  solar radiation with artificial intelligence techniques, {\em Solar Energy}
  {\bf 68}(2)  (2000)  169--178.

\bibitem{CHANDRA:2012}
R.~Chandra and M.~Zhang, Cooperative coevolution of {Elman} recurrent neural
  networks for chaotic time series prediction, {\em Neurocomputing} {\bf 86}
  (2012)  116 -- 123.

\bibitem{Mohammadi:2018}
M.~Mohammadi, F.~Talebpour, E.~Safaee, N.~Ghadimi and O.~Abedinia, Small-scale
  building load forecast based on hybrid forecast engine, {\em Neural
  Processing Letters} {\bf 48}(1)  (2018)  329--351.

\bibitem{Ruiz:2018}
L.~Ruiz, R.~Rueda, M.~Cuéllar and M.~Pegalajar, Energy consumption forecasting
  based on {{Elman}} neural networks with evolutive optimization, {\em Expert
  Syst. Appl.} {\bf 92}  (2018)  380 -- 389.

\bibitem{Schafer:2006}
A.~M. Sch\"{a}fer and H.~G. Zimmermann, Recurrent neural networks are universal
  approximators, {\em Proc. 16th ICANN\/},  (Springer-Verlag, 2006), pp.
  632--640.

\bibitem{Elman:1990}
J.~L. {Elman}, Finding structure in time, {\em Cognitive Science} {\bf 14}(2)
  (1990)  179 -- 211.

\bibitem{Kim:2018}
H.~Kim and C.~Won, Forecasting the volatility of stock price index: A hybrid
  model integrating {LSTM} with multiple {GARCH}-type models, {\em Expert Syst.
  Appl.} {\bf 103}  (2018)  25--37.

\bibitem{Yu:2018}
C.~Yu, Y.~Li, Y.~Bao, H.~Tang and G.~Zhai, A novel framework for wind speed
  prediction based on recurrent neural networks and support vector machine,
  {\em Energy Convers. Manag.} {\bf 178}  (2018)  137--145.

\bibitem{Wang:2019}
Y.~Wang, Y.~Shen, S.~Mao, X.~Chen and H.~Zou, {LASSO} and {LSTM} integrated
  temporal model for short-term solar intensity forecasting, {\em IEEE Internet
  Things} {\bf 6}(2)  (2019)  2933--2944.

\bibitem{M4:2018}
S.~Makridakis, E.~Spiliotis and V.~Assimakopoulos, The {M4} competition:
  {Results}, findings, conclusion and way forward, {\em International Journal
  of Forecasting} {\bf 34}(4)  (2018)  802 -- 808.

\bibitem{Doya:1989}
K.~Doya and S.~Yoshizawa, Adaptive neural oscillator using continuous-time
  back-propagation learning, {\em Neural Networks} {\bf 2}(5)  (1989)
  375--385.

\bibitem{Jordan:1986}
M.~I. Jordan, {Chapter 25 - Serial Order: A Parallel Distributed Processing
  Approach}, {\em Institute for Cognitive Science Technical Report 8604\/},
  1986.

\bibitem{Bengio:1994}
Y.~{Bengio}, P.~{Simard} and P.~{Frasconi}, Learning long-term dependencies
  with gradient descent is difficult, {\em IEEE Trans. Neural Netw.} {\bf 5}(2)
   (1994)  157--166.

\bibitem{Ma:2015}
X.~Ma, Z.~Tao, Y.~Wang, H.~Yu and Y.~Wang, Long short-term memory neural
  network for traffic speed prediction using remote microwave sensor data, {\em
  Transportation Research Part C: Emerging Technologies} {\bf 54}  (2015)
  187--197.

\bibitem{Tian:2015}
Y.~Tian and L.~Pan, Predicting short-term traffic flow by long short-term
  memory recurrent neural network. {\em Proceedings IEEE ISC2 2015} , pp.
  153--158.

\bibitem{Fischer:2018}
T.~Fischer and C.~Krauss, Deep learning with long short-term memory networks
  for financial market predictions, {\em European Journal of Operational
  Research} {\bf 270}(2)  (2018)  654--669.

\bibitem{Bouktif:2018}
S.~Bouktif, A.~Fiaz, A.~Ouni and M.~Serhani, Optimal deep learning {LSTM} model
  for electric load forecasting using feature selection and genetic algorithm:
  Comparison with machine learning approaches, {\em Energies} {\bf 11}(7)
  (2018).

\bibitem{Sagheer:2019}
A.~Sagheer and M.~Kotb, Time series forecasting of petroleum production using
  deep {LSTM} recurrent networks, {\em Neurocomputing} {\bf 323}  (2019)  203
  -- 213.

\bibitem{tan:2019}
C.~Pan, J.~Tan, D.~Feng and Y.~Li, Very short-term solar generation forecasting
  based on {LSTM} with temporal attention mechanism, {\em 2019 IEEE 5th
  ICCC\/},  pp. 267--271.

\bibitem{Smyl:2020}
S.~Smyl, A hybrid method of exponential smoothing and recurrent neural networks
  for time series forecasting, {\em Int. J. Forecast.} {\bf 36}(1)  (2020)  75
  -- 85.

\bibitem{Bandara:2020}
K.~Bandara, C.~Bergmeir and S.~Smyl, Forecasting across time series databases
  using recurrent neural networks on groups of similar series: A clustering
  approach, {\em Expert Syst. Appl.} {\bf 140}  (2020) p. 112896.

\bibitem{Lin:2009}
X.~Lin, Z.~Yang and Y.~Song, Short-term stock price prediction based on echo
  state networks, {\em Expert Syst. Appl.} {\bf 36}(3)  (2009)  7313--7317.

\bibitem{Rodan:2011}
A.~Rodan and P.~Tiňo, Minimum complexity echo state network, {\em IEEE Trans.
  Neural Netw.} {\bf 22}(1)  (2011)  131--144.

\bibitem{Li:2012}
D.~Li, M.~Han and J.~Wang, Chaotic time series prediction based on a novel
  robust echo state network, {\em IEEE Trans. Neural Netw. Learn. Syst.} {\bf
  23}(5)  (2012)  787--797.

\bibitem{Deihimi:2012}
A.~Deihimi and H.~Showkati, Application of echo state networks in short-term
  electric load forecasting, {\em Energy} {\bf 39}(1)  (2012)  327--340.

\bibitem{Liu:2015}
D.~Liu, J.~Wang and H.~Wang, Short-term wind speed forecasting based on
  spectral clustering and optimised echo state networks, {\em Renew. Energ.}
  {\bf 78}  (2015)  599--608.

\bibitem{Hu:2020}
H.~Hu, L.~Wang, L.~Peng and Y.~Zeng, Effective energy consumption forecasting
  using enhanced bagged echo state network, {\em Energy} {\bf 193}  (2020).

\bibitem{LSTM:1997}
S.~Hochreiter and J.~Schmidhuber, Long short-term memory, {\em Neural
  computation} {\bf 9}  (1997)  1735--80.

\bibitem{Gers:2000}
F.~A. Gers, J.~Schmidhuber and F.~Cummins, Learning to forget: Continual
  prediction with {LSTM}, {\em Neural Computation} {\bf 12}(10)  (2000)
  2451--2471.

\bibitem{Chung:2014}
J.~Chung, {\c{C}}.~G{\"{u}}l{\c{c}}ehre, K.~Cho and Y.~Bengio, Empirical
  evaluation of gated recurrent neural networks on sequence modelling, {\em
  CoRR} {\bf abs/1412.3555}  (2014).

\bibitem{Kuan:2017}
L.~Kuan, Z.~Yan, W.~Xin, C.~Yan, P.~Xiangkun, S.~Wenxue, J.~Zhe, Z.~Yong,
  X.~Nan and Z.~Xin, Short-term electricity load forecasting method based on
  multilayered self-normalizing {GRU} network. {\em 2017 IEEE Conference EI2} ,
  pp. 1--5.

\bibitem{WangLiao:2018}
Y.~Wang, W.~Liao and Y.~Chang, Gated recurrent unit network-based short-term
  photovoltaic forecasting, {\em Energies} {\bf 11}(8)  (2018).

\bibitem{Ugurlu:2018}
U.~Ugurlu, I.~Oksuz and O.~Tas, Electricity price forecasting using recurrent
  neural networks, {\em Energies} {\bf 11}(5)  (2018).

\bibitem{ESN:2001}
H.~Jaeger, The echo state approach to analysing and training recurrent neural
  networks-with an erratum note, {\em German National Research Center for
  Information Technology Report} {\bf 148}  (2001).

\bibitem{IsmailFawaz:2018}
H.~Ismail~Fawaz, G.~Forestier, J.~Weber, L.~Idoumghar and P.-A. Muller, Deep
  learning for time series classification: a review, {\em Data Mining and
  Knowledge Discovery} {\bf 33}(4)  (2019)  917--963.

\bibitem{Jaeger:2004}
H.~Jaeger and H.~Haas, Harnessing nonlinearity: Predicting chaotic systems and
  saving energy in wireless communication, {\em Science} {\bf 304}(5667)
  (2004)  78--80.

\bibitem{GRU:2014}
K.~Cho, B.~van Merrienboer, C.~Gulcehre, F.~Bougares, H.~Schwenk and Y.~Bengio,
  Learning phrase representations using {RNN} encoder{--}decoder for
  statistical machine translation, {\em Proc. EMNLP 2014\/},  pp. 1724--1734.

\bibitem{Ravanelli:2018}
M.~Ravanelli, P.~Brakel, M.~Omologo and Y.~Bengio, Light gated recurrent units
  for speech recognition, {\em IEEE T. Emerg. Top. Com.} {\bf 2}(2)  (2018)
  92--102.

\bibitem{Tsantekidis:2017}
A.~{Tsantekidis}, N.~{Passalis}, A.~{T}, J.~{K}, M.~{Gabbouj} and
  A.~{Iosifidis}, Forecasting stock prices from the limit order book using
  convolutional neural networks, {\em 2017 IEEE CBI\/},   {\bf 01}, pp. 7--12.

\bibitem{Kuo:2018}
P.-H. Kuo and C.-J. Huang, A high precision artificial neural networks model
  for short-term energy load forecasting, {\em Energies} {\bf 11}(1)  (2018).

\bibitem{Koprinska:2018}
I.~{Koprinska}, D.~{Wu} and Z.~{Wang}, Convolutional neural networks for energy
  time series forecasting, {\em IJCNN 2018\/},  pp. 1--8.

\bibitem{Tian:2018}
C.~Tian, J.~Ma, C.~Zhang and P.~Zhan, A deep neural network model for
  short-term load forecast based on long short-term memory network and
  convolutional neural network, {\em Energies} {\bf 11}(12)  (2018).

\bibitem{Liu:2018}
H.~Liu and Y.~Li, Smart deep learning based wind speed prediction model using
  wavelet packet decomposition, convolutional neural network and convolutional
  long short term memory network, {\em Energy Conversion and Management} {\bf
  166}  (2018)  120 -- 131.

\bibitem{Cai:2019}
M.~Cai, M.~Pipattanasomporn and S.~Rahman, Day-ahead building-level load
  forecasts using deep learning vs. traditional time-series techniques, {\em
  Applied Energy} {\bf 236}  (2019)  1078 -- 1088.

\bibitem{Shen:2019}
Z.~Shen, Y.~Zhang, J.~Lu, J.~Xu and G.~Xiao, A novel time series forecasting
  model with deep learning, {\em Neurocomputing} {\bf 396}  (2020)  302 -- 313.

\bibitem{Borovykh:2019}
A.~Borovykh, S.~Bohte and C.~Oosterlee, Dilated convolutional neural networks
  for time series forecasting, {\em Journal of Computational Finance} {\bf
  22}(4)  (2019)  73--101.

\bibitem{Chen:2019}
Y.~Chen, Y.~Kang, Y.~Chen and Z.~Wang, Probabilistic forecasting with temporal
  convolutional neural network, {\em arXiv:1906.04397}   (2019).

\bibitem{Wan:2019}
R.~Wan, S.~Mei, J.~Wang, M.~Liu and F.~Yang, Multivariate temporal
  convolutional network: A deep neural networks approach for multivariate time
  series forecasting, {\em Electronics} {\bf 8}(8)  (2019).

\bibitem{Lara-Benitez:2020}
P.~Lara-Ben{\'{i}}tez, M.~Carranza-Garc{\'{i}}a, J.~Luna-Romera and
  J.~Riquelme, {Temporal Convolutional Networks Applied to Energy-Related Time
  Series Forecasting}, {\em Applied Sciences} {\bf 10}(7)  (2020) p. 2322.

\bibitem{Kang:2018}
D.~Kang and Y.-J. Cha, Autonomous {UAVs} for structural health monitoring using
  deep learning and an ultrasonic beacon system with geo-tagging, {\em
  Computer-Aided Civil and Infrastructure Engineering} {\bf 33}(10)  (2018)
  885--902.

\bibitem{Hinton:2012}
G.~{Hinton}, L.~{Deng}, D.~{Yu}, G.~E. {Dahl}, A.~{Mohamed}, N.~{Jaitly},
  A.~{Senior}, V.~{Vanhoucke}, P.~{Nguyen}, T.~N. {Sainath} and B.~{Kingsbury},
  Deep neural networks for acoustic modeling in speech recognition: The shared
  views of four research groups, {\em IEEE Signal Processing Magazine} {\bf
  29}(6)  (2012)  82--97.

\bibitem{Kim:2014}
Y.~Kim, Convolutional neural networks for sentence classification, {\em
  Proceedings EMNLP 2014 Conference\/},  pp. 1746--1751.

\bibitem{yang:2015}
J.-B. Yang, N.~Nhut, P.~San, X.~li and P.~Shonali, Deep convolutional neural
  networks on multichannel time series for human activity recognition, {\em
  IJCAI}  (07 2015).

\bibitem{Oord:2016}
A.~Oord, S.~Dieleman, H.~Zen, K.~Simonyan, O.~Vinyals, A.~Graves,
  N.~Kalchbrenner, A.~Senior and K.~Kavukcuoglu, Wavenet: A generative model
  for raw audio, {\em arXiv:11609.03499}   (2016).

\bibitem{Bai:2018}
S.~Bai, J.~Kolter and V.~Koltun, An empirical evaluation of generic
  convolutional and recurrent networks for sequence modeling, {\em
  arXiv:1803.01271}   (2018).

\bibitem{Yu:2016}
F.~Yu and V.~Koltun, Multi-scale context aggregation by dilated convolutions,
  {\em {ICLR} 2016\/},

\bibitem{github-code}
P.~Lara-Benítez and M.~Carranza-García, {Time Series Forecasting - Deep
  Learning}
  \url{https://github.com/pedrolarben/TimeSeriesForecasting-DeepLearning},
  (2020).

\bibitem{CIF:data}
M.~Štěpnička and M.~Burda, {C}omputational {I}ntelligence in {F}orecasting
  ({CIF}) \url{https://irafm.osu.cz/cif},  (2016).

\bibitem{ExchangeRate:data}
G.~Lai, W.-C. Chang, Y.~Yang and H.~Liu, Modeling long- and short-term temporal
  patterns with deep neural networks, {\em arXiv:1703.07015}   (2017).

\bibitem{M3:data:paper}
S.~Makridakis and M.~Hibon, The {M3}-competition: results, conclusions and
  implications, {\em International Journal of Forecasting} {\bf 16}(4)  (2000)
  451 -- 476.

\bibitem{M4:data:paper}
S.~Makridakis, E.~Spiliotis and V.~Assimakopoulos, The {M4} competition:
  100,000 time series and 61 forecasting methods, {\em International Journal of
  Forecasting} {\bf 36}(1)  (2020)  54 -- 74.

\bibitem{NN5:2008}
NNGC, {NN5} time series forecasting competition for neural networks
  \url{http://www.neural-forecasting-competition.com/NN5},  (2008).

\bibitem{Tourism:data}
G.~Athanasopoulos, R.~J. Hyndman, H.~Song and D.~C. Wu, Tourism forecasting
  part two \url{www.kaggle.com/c/tourism2},  (2010).

\bibitem{SolarEnergy:data}
NREL, Solar power data for integration studies
  \url{www.nrel.gov/grid/solar-power-data.html},  (2007).

\bibitem{Traffic-la-bay:data}
Y.~Li, R.~Yu, C.~Shahabi and Y.~Liu, Diffusion convolutional recurrent neural
  network: Data-driven traffic forecasting, {\em arXiv:1707.01926}   (2017).

\bibitem{WikiWebTraffic:data}
Google, Web traffic time series forecasting competition
  \url{www.kaggle.com/c/web-traffic-time-series-forecasting},  (2017).

\bibitem{Bentaieb:2012}
S.~{Ben Taieb}, G.~Bontempi, A.~F. Atiya and A.~Sorjamaa, A review and
  comparison of strategies for multi-step ahead time series forecasting based
  on the {NN5} forecasting competition, {\em Expert Syst. Appl.} {\bf 39}(8)
  (2012)  7067 -- 7083.

\bibitem{Kingma:2014}
D.~Kingma and J.~Ba, Adam: A method for stochastic optimization, {\em
  arXiv:1412.6980}   (2014).

\bibitem{KIM:2016}
S.~Kim and H.~Kim, A new metric of absolute percentage error for intermittent
  demand forecasts, {\em Int. J. Forecast} {\bf 32}(3)  (2016)  669 -- 679.

\bibitem{Garcia:2008}
S.~García and F.~Herrera, An extension on ``statistical comparisons of
  classifiers over multiple data sets" for all pairwise comparisons, {\em
  Journal of Machine Learning Research} {\bf 9}  (2008)  2677--2694.

\bibitem{Wilcoxon:1992}
F.~Wilcoxon, {\em Individual Comparisons by Ranking Methods}, {\em
  Breakthroughs in Statistics: Methodology and Distribution\/} (Springer,
  1992), pp. 196--202.

\end{thebibliography}
\end{multicols}
\end{document}